\documentclass[12pt,letterpaper]{article}
\usepackage[a4paper, total={7in, 10in}]{geometry}

\usepackage{helvet}
\usepackage{authblk}
\usepackage{hyperref}
\usepackage{graphicx}
\usepackage{todonotes}
\usepackage{cleveref}
\usepackage{import}
\usepackage[subpreambles=false]{standalone}
\usepackage{setspace}
\usepackage{verbatim}
\usepackage[super,comma,sort&compress]{natbib}
\usepackage{xurl}
\usepackage{pdfpages}

\newcommand{\numpy}{\texttt{NumPy}}
\newcommand{\acia}{\texttt{acia}}
\newcommand{\aw}{\texttt{acia-workflows}}

\newcommand{\SegmentationProcessor}{\texttt{SegmentationProcessor}}
\newcommand{\ImageSequenceSource}{\texttt{ImageSequenceSource}}
\newcommand{\Overlay}{\texttt{Overlay}}
\newcommand{\TrackingProcessor}{\texttt{TrackingProcessor}}
\newcommand{\networkx}{\texttt{networkx}}
\newcommand{\pandas}{\texttt{Pandas}}
\newcommand{\DiGraph}{\texttt{DiGraph}}
\newcommand{\LAP}{\texttt{LAP}}
\newcommand{\trackastra}{\texttt{Trackastra}}
\newcommand{\PyUAT}{\texttt{PyUAT}}
\newcommand{\pint}{\texttt{pint}}
\newcommand{\moviepy}{\texttt{moviepy}}
\newcommand{\papermill}{\texttt{papermill}}
\newcommand{\CG}{\textit{C.~glutamicum}}
\newcommand{\EC}{\textit{E.~coli}}
\newcommand{\mvenus}{mVenus}
\newcommand{\crimson}{E2-Crimson}
\newcommand{\scipy}{\texttt{scipy}}

\newcommand{\sixkey}{ACMS2R}
\newcommand{\SO}{SOTA}

\makeatletter
\renewcommand{\maketitle}{\bgroup\setlength{\parindent}{0pt}
\begin{flushleft}
\par\vspace{5pt}
  \textbf{\Large \@title}
\par\vspace{5pt}
  \@author
\end{flushleft}\egroup}
\makeatother

\title{acia-workflows: Automated Single-cell Imaging Analysis for Scalable and Deep Learning-based Live-cell Imaging Analysis Workflows}
\date{}

\author[1,2]{Johannes Seiffarth}
\author[1,2]{Keitaro Kasahara}
\author[1]{Michelle Bund}
\author[1]{Benita Lückel}
\author[3,4]{Richard D. Paul}
\author[1,2]{Matthias Pesch}
\author[1,2]{Lennart Witting}
\author[1,5]{Michael Bott}
\author[1]{Dietrich Kohlheyer}
\author[1,*]{Katharina Nöh}

\affil[1]{\small Institute for Bio- and Geosciences, IBG-1: Biotechnology, Forschungszentrum Jülich, Jülich, Germany}
\affil[2]{\small Computational Systems Biotechnology (AVT.CSB), RWTH Aachen University, Aachen, Germany}
\affil[3]{\small Institute for Advanced Simulation, IAS-8: Data Analytics and Machine Learning, Forschungszentrum Jülich, Jülich, Germany}
\affil[4]{\small Department for Statistics, Ludwig Maximilian University of Munich, Munich, Germany}
\affil[5]{\small Bioeconomy Science Center (BioSC), Forschungszentrum Jülich, Jülich, Germany}

\affil[*]{\small Correspondence: \href{mailto:k.noeh@fz-juelich.de}{k.noeh@fz-juelich.de}}

\begin{document}

\maketitle

\section*{Summary}

Live-cell imaging (LCI) technology enables the detailed spatio-temporal characterization of living cells at the single-cell level, which is critical for advancing research in the life sciences, from biomedical applications to bioprocessing. High-throughput setups with tens to hundreds of parallel cell cultivations offer the potential for robust and reproducible insights. However, these insights are obscured by the large amount of LCI data recorded per experiment. Recent advances in state-of-the-art deep learning methods for cell segmentation and tracking now enable the automated analysis of such large data volumes, offering unprecedented opportunities to systematically study single-cell dynamics. The next key challenge lies in integrating these powerful tools into accessible, flexible, and user-friendly workflows that support routine application in biological research. In this work, we present \aw{}, a platform that combines three key components: (1) the Automated live-Cell Imaging Analysis (\acia) Python library, which supports the modular design of image analysis pipelines offering eight deep learning segmentation and tracking approaches; (2) workflows that assemble the image analysis pipeline, its software dependencies, documentation, and visualizations into a single Jupyter Notebook, leading to accessible, reproducible and scalable analysis workflows; and (3) a collection of application workflows showcasing the analysis and customization capabilities in real-world applications. Specifically, we present three workflows to investigate various types of microfluidic LCI experiments ranging from growth rate comparisons to precise, minute-resolution quantitative analysis of individual dynamic cells responses to changing oxygen conditions. Our collection of more than ten application workflows is open source and publicly available at \url{https://github.com/JuBiotech/acia-workflows}.

\section*{Introduction}
Live-cell imaging is at the forefront of investigating the dynamic behavior of living cells across space and time, fostering our understanding of cancer treatment\cite{alieva_bridging_2023}, protein secretion\cite{shirasaki_real-time_2014,raphael_quantitative_2013}, diseases\cite{weissmann_mechanisms_2008,campbell_live_2008}, single-cell heterogeneity~\cite{preedy_cellular_2024,huang_exploring_2024}, and biofilm formation\cite{drescher_solutions_2014,hartmann_emergence_2019}. Combining automated live-cell imaging with disposable, high-throughput microfluidic devices (MLCI) enables the phenotyping of single cells and their development into cell populations in precisely controlled environments~\cite{cornaglia_microfluidic_2017,ortseifen_microfluidics_2020,weibel_microfabrication_2007}. These MLCI setups simultaneously record numerous cell populations within a single experimental run at constant or time-varying conditions. Their picoliter scale and high-throughput nature promise resource-efficient yet reliable single-cell insights by increasing sample size while reducing the number of experiments. 
Thus, MLCI is ideally suited for studying dynamic biological processes that occur over short timescales, delineating the growth behavior of single cells over many generations\cite{witting_microfluidic_2025,kasahara_unveiling_2025, tauber_dmscc_2020}, and the interplay of cell-to-cell interactions\cite{schito_communities_2022,burmeister_microfluidic_2020}, including the emergence of single-cell heterogeneity \cite{chung_single-cell_2024}. 

The key to the power of MLCI lies in extracting single-cell information from time-lapse data and analyzing the spatio-temporal development of cells and their populations. However, these single-cell measurements are hidden within the imaging data and must be extracted from tens to hundreds of gigabytes of time-lapse data. This extraction requires highly automated image analysis pipelines.

\begin{figure}
    \centering
    \includegraphics[width=0.95\linewidth]{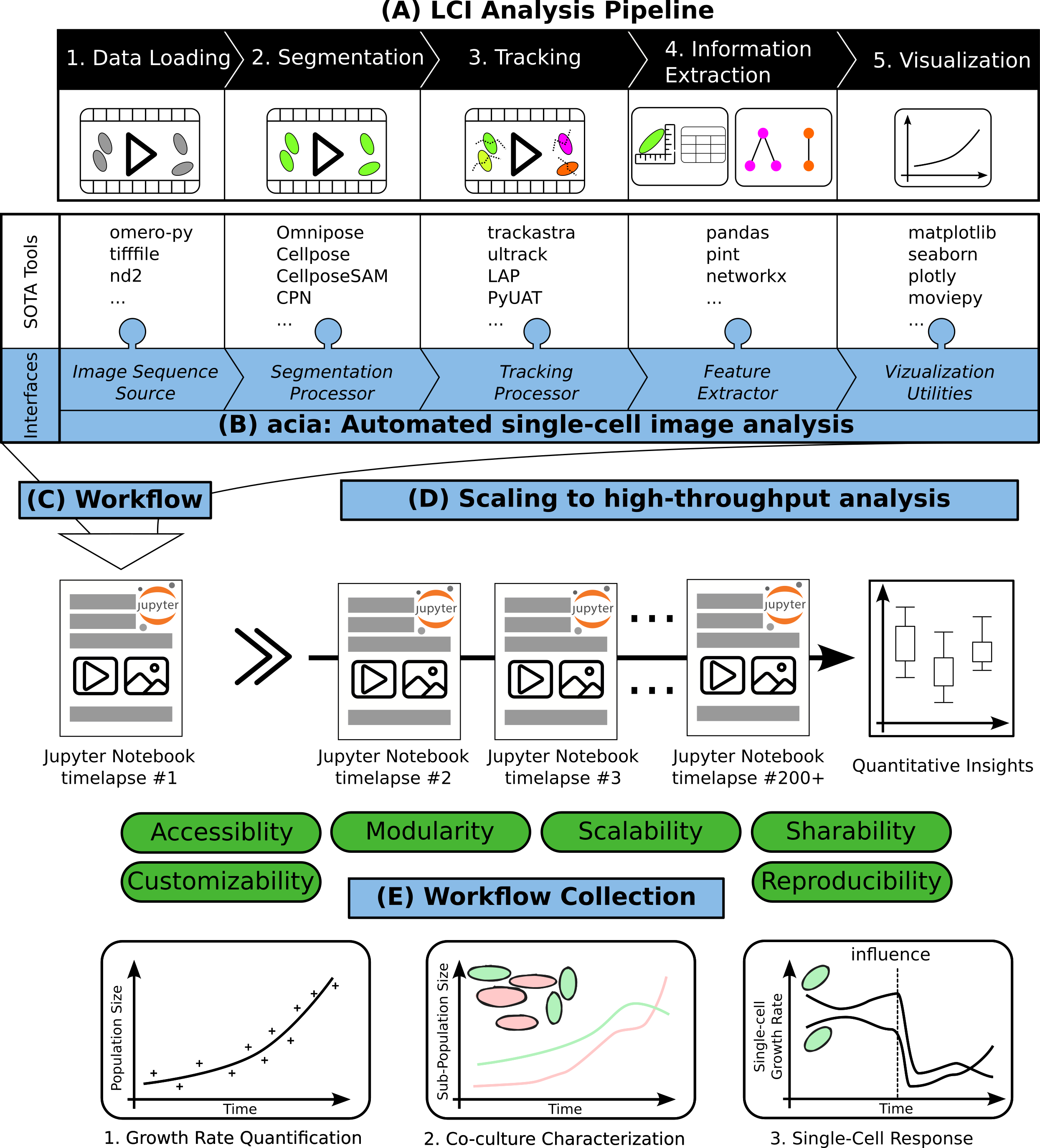}
    \caption{Five-step MLCI analysis pipeline (A), implemented in modular components of the \acia{} library (B) utilizing \SO{} methods and existing Python libraries. These steps are implemented sequentially within a single Jupyter Notebook, fusing code, documentation, software dependencies, and visualizations into a single workflow (C). This workflow is automatically scaled to high-throughput experiments with numerous time-lapse recordings to gain quantitative insights (D). Our workflow collection (E) showcases the importance of the six key capabilities: \textit{accessibility}, \textit{customizability}, \textit{modularity}, \textit{scalability}, \textit{shareability}, and \textit{reproducibility} (\sixkey).}
    \label{fig:analysis_workflow}
\end{figure}

\Cref{fig:analysis_workflow}A shows the typical five steps of such an MLCI image analysis pipeline: First, the time-lapse imaging data is loaded, then cell objects are detected within every microscopy image using instance segmentation. These cell objects are tracked across images to reconstruct their temporal context, followed by extracting the biological single-cell features such as cell size, fluorescence, or age. Finally, the extracted single-cell features are used to derive and visualize live-cell insights.

For a long time, the main challenge for image analysis pipelines has been developing tools for each step of the pipeline. In particular, the segmentation and tracking of living cells in time-lapses has proven to be challenging~\cite{jeckel_advances_2021,yazdi_survey_2024}. With the development of a plethora of modern deep learning (DL) tools for segmentation~\cite{cutler_omnipose_2022,stringer_cellpose_2021,upschulte_uncertainty-aware_2023,Seiffarth2025,scherr_microbeseg_2022} and tracking~\cite{gallusser_trackastra_2024,jaqaman_robust_2008,oconnor_delta_2022,OConnor2025} that benefit from emerging datasets~\cite{Seiffarth2025,stringer_cellpose_2021,schwartz_caliban_2023,edlund_livecelllarge-scale_2021}, full automation of these two steps is now within reach. 
Consequently, the central challenge of the analysis is shifting: Rather than optimizing and automating individual steps separately, the challenge shifts to integrating them into a coherent and robust pipeline and making them available to a broad community of life scientists. 
Based on our experience of deploying MLCI image analysis pipelines,
we have identified the six most important capabilities for such image analysis pipelines: \textit{accessibility}, \textit{customizability}, \textit{modularity}, \textit{scalability}, \textit{shareability}, and \textit{reproducibility} (\sixkey).

These ability requirements arise from the sequential pipeline steps and general challenges in deploying software in a research lab.
Time-lapse data originates from diverse microscopy platforms and vendor software, resulting in heterogeneous formats. These datasets include imaging channels (e.g., phase contrast, bright field, fluorescence) and metadata. To handle such variability, flexible \textit{customization} of data and metadata loading is essential.

To benefit from the latest advances in DL methods for cell segmentation and tracking in practice, users must be able to quickly test state-of-the-art (SOTA) methods, tune their parameters, and integrate new methods with ease. Thus, \textit{modularity} with clear interfaces is crucial.
Segmentation and tracking yield single-cell data that form the basis of biological insights. However, research questions vary widely—ranging from interest in cell size, fluorescence markers, or abnormal growth to cell death or dynamic behaviors. Hence, \textit{customizable} feature extraction and visualization are vital to match evolving research needs.

Once such a pipeline automatically extracts single-cell insights, it is critical to quantify the variability of the measured effects and patterns of single-cell behavior. Consequently, such experiments are typically performed in a high-throughput manner to capture data from multiple independent cell populations. Thus, it is important for users to quickly roll out the pipeline across multiple replicates (\textit{scalability}) and gather the information to quantify the variability in the extracted insights.

Additionally, three more capabilities arise from deploying such pipelines in our research lab, all of which boost productivity: \textit{accessibility}, \textit{sharability}, and \textit{reproducibility}. We define good \textit{accessibility} regarding low-entry barriers for users with little to no programming experience and eliminating complex software installation procedures or specific hardware requirements. The entry to the analysis software should be as easy as opening a website in the browser to democratize its usage among a broad research community. Moreover, the usage of such software is boosted by the ability to \textit{share} and \textit{reproduce} analysis results among teams and fellow researchers. Both facilitate the verification of the extracted insights and allow for continuous improvement and extension of the analysis pipeline, turning the development and improvement of such analysis pipelines into a joint community effort.

In recent years, numerous analysis pipelines have been developed that demonstrate the enormous potential of automated image analysis\cite{oconnor_delta_2022,berg_ilastik_2019,stirling_cellprofiler_2021,stylianidou_supersegger_2016,Lo2025,ouyang_imjoy_2019,luik_biomero_2024}. However, these pipelines are usually centered around a specific segmentation or tracking approach, limiting their \textit{modularity} and making their application across various imaging modalities and cell morphologies challenging. In contrast, plugin-based analysis tools such as Fiji~\cite{schindelin_fiji_2012} and napari~\cite{chiu_napari_2022} provide a wide range of segmentation and tracking approaches \cite{ershov_trackmate_2022,gallusser_trackastra_2024,ducret_microbej_2016,ouyang_bioimage_2022} and offer tremendous visualization capabilities. However, users need to assemble their own pipelines from scratch, which makes these tools especially useful for advanced users, but lowers the \textit{accessibility} for beginners.

To overcome the limitations of existing pipelines in addressing the six key capabilities, we present \aw{} - a time-lapse analysis platform that integrates \SO{} tools into \sixkey{} workflows. To achieve this, \aw{} combines three complementary components: First, the \acia{} Python library implements the modular time-lapse analysis pipeline (\Cref{fig:analysis_workflow}B). Second, a workflow concept that allows to integrate code, documentation, and visualizations into a single traceable document (\Cref{fig:analysis_workflow}C), and third, an open-source collection of application workflows.

The \acia{} Python library implements \textit{modular} interfaces (\Cref{fig:analysis_workflow}B), integrating eight \SO{} segmentation and tracking methods and chains all pipeline steps into one sequential execution. Modules defined by interfaces can be quickly swapped in and out, making changes in the segmentation or tracking methods convenient. We implement this sequential pipeline into Jupyter Notebooks, fusing the Python code with rich documentation, visualization, and all software dependencies into a single document that we term a "workflow" (\Cref{fig:analysis_workflow}C). Such a Jupyter Notebook-based workflow contains the complete image analysis workflow within a single document and makes it \textit{accessible}, \textit{reproducible}, and \textit{customizable} in the web browser~\cite{kluyver_jupyter_2016,von_chamier_democratising_2021} and high-performance computing systems\cite{von_chamier_democratising_2021}.
We specifically design the workflows to be easily applied to multiple time-lapse sequences, thereby achieving high \textit{scalability} and unlocking quantitative insights across multiple replicates (\Cref{fig:analysis_workflow}D).

Finally, we demonstrate the impact of its six key capabilities at the example of three MLCI analysis workflows: 
(1) Quantifying population growth rates, 
(2) performing co-culture characterization, and 
(3) measuring single-cell responses to changes in cultivation conditions. 
We utilize the scalability of our workflows and apply them to multiple replicates without requiring manual code changes. To emphasize \textit{accessibility}, \textit{sharability}, and \textit{reproducibility}, the workflows are available open-source along with a comprehensive set of over 10 application workflows.  These can be reproduced with GPU acceleration directly in the web browser using Google Colab.

\section*{Image analysis pipeline and workflow}

The image analysis of MLCI time-lapses requires a multi-step pipeline where different modules deal with image formats, segmentation, tracking, cell feature extraction, and visualization or insight generation (\Cref{fig:analysis_workflow}A).
Thus, we develop these modules within the \acia{} Python library with the \sixkey{} abilities in mind and chain them sequentially into a so-called "workflow". This workflow is implemented within a Jupyter Notebook combining software dependencies, documentation, visualizations, video renderings, and custom code into one traceable document~\cite{von_chamier_democratising_2021}. We introduce the capability to \textit{scale} these workflows to numerous time-lapses without manual code changes.

\subsection*{Loading 2D+t microscopy time-lapse data}

Live-cell microscopy time-lapse data comes in various formats: the popular bio-formats framework supports up to $160$ different image and metadata formats (\url{http://www.openmicroscopy.org/bio-formats/}). Handling and processing all these different formats is challenging and requires a \textit{customizable} data loading procedure. Thus, we define a standard data interface for 2D+t live-cell time-lapse data represented in a $T \times H \times W \times C$ shaped vector, denoting the time ($T$), spatial imaging ($H, W$), and channel ($C$) dimensions. To process the data, we implement the interface \texttt{ImageSequenceSource} providing an iterator along the temporal dimension yielding $H \times W \times C$ images as \numpy{} arrays\cite{harris_array_2020}. The user provides a short \textit{custom} code snippet to load the specific imaging format and convert it into a $T \times H \times W \times C$ \numpy{} array. In addition, we provide an implementation to stream image stacks from OMERO servers~\cite{allan_omero_2012}. Thus, \acia breaks down loading 2D+t time-lapse data to a few lines of Python code, regardless of whether they are stored locally, in an online OMERO database, or \textit{custom} data format.

\subsection*{Cell segmentation}

The first step in extracting single-cell information from 2D+t sequences is segmentation, i.e., the pixel-precise detection of the individual cells (\Cref{fig:analysis_workflow}A). Numerous DL approaches have been developed in recent years, including \texttt{Cellpose}\cite{stringer_cellpose_2021}, \texttt{CellposeSAM}\cite{pachitariu_cellpose-sam_2025}, \texttt{Omnipose}\cite{cutler_omnipose_2022}, and \texttt{CPN}\cite{upschulte_uncertainty-aware_2023}, showing excellent performance for a broad range of cell morphologies and imaging modalities. To transfer this high-quality segmentation into everyday usage, \acia{} \textit{modularly} integrates these \SO{} approaches. Therefore, we define a \SegmentationProcessor{} interface in Python that receives a 2D+t time-lapse (\ImageSequenceSource) and computes a segmentation overlay (\Overlay) containing cell instances, represented as cell masks or cell contours in polygon format (\Cref{fig:feature_formats}B). We provide implementations for \texttt{Cellpose}, \texttt{CellposeSAM}, \texttt{Omnipose}, and \texttt{CPN} to allow users to quickly apply different segmentation models and select the most suitable method for their cell morphology and imaging modality. The generic interface fosters future cell segmentation methods to be integrated into the pipeline.

\subsection*{Cell tracking}

As for cell segmentation, numerous cell tracking approaches have been developed~\cite{gallusser_trackastra_2024,jaqaman_robust_2008,oconnor_delta_2022,Seiffarth2025, paul2025,bragantini_large-scale_2025}. Most common is the tracking-by-detection paradigm, where first cell segments are detected in the images that are then linked through time. Thus, \acia{} defines the \TrackingProcessor{} interface that uses the 2D+t image sequence (\ImageSequenceSource) and the segmentation information (\Overlay) and computes the resulting tracking lineage as a graph structure (\networkx{} \DiGraph{}, \url{https://github.com/networkx/networkx}). In this graph structure, every detected cell instance is represented as a node, and edges link them through time (\Cref{fig:feature_formats}C). These linked cell detections form "tracklets", which cover the entire cell cycle - from cell birth to division or disappearance. These tracklets are given unique labels and stored in a tracklet graph. In this graph, a node represents the entire cell cycle, and edges indicate cell divisions (\Cref{fig:feature_formats}D).
We \textit{modularly} implement the interface for four tracking methods including classic \LAP~\cite{jaqaman_robust_2008}, transformer-based \trackastra~\cite{gallusser_trackastra_2024}, a Bayesian cell tracker \PyUAT~\cite{Seiffarth_2025}, and a multi-segmentation hypothesis tracker \texttt{Ultrack}\cite{bragantini_large-scale_2025}. Similar to segmentation models, adding custom tracking implementations is straightforward.

\subsection*{Unit-aware spatio-temporal cell feature extraction}

\begin{figure}
    \centering
    \includegraphics[width=1.0\linewidth]{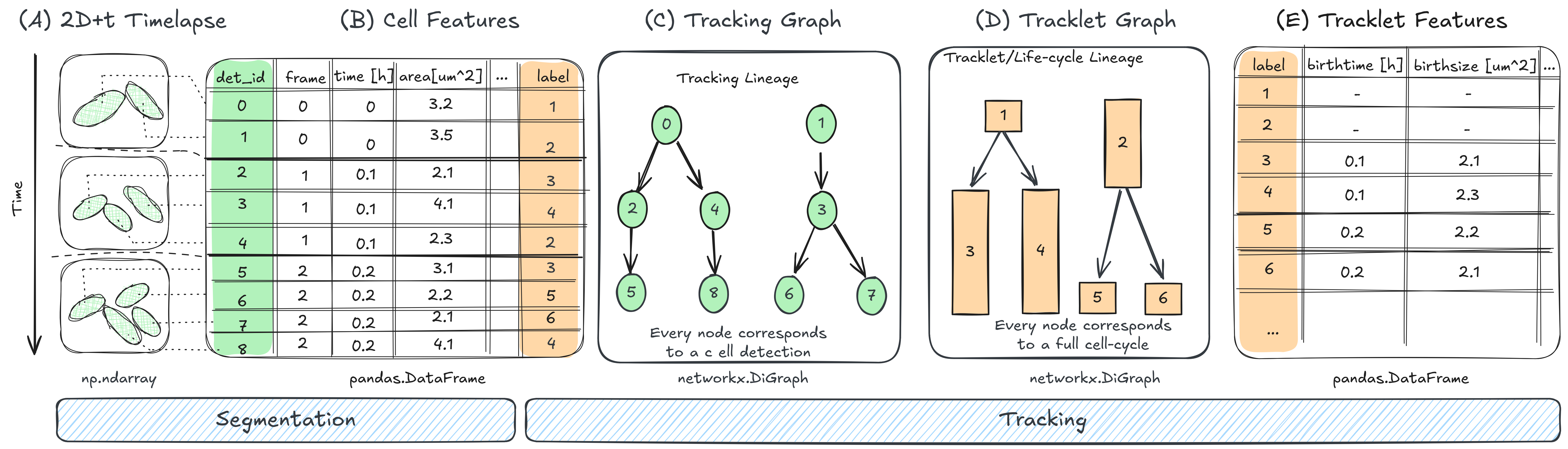}
    \caption{\textbf{Data extraction within \aw.} \textbf{A:} The 2D+t time-lapse is segmented to extract cell instances. \textbf{B:} Spatial single-cell features are extracted with units. \textbf{C \& D:} Cell instances are associated through time to build the tracking and tracklet lineages. \textbf{E:} Temporal single-cell features are extracted with units and linked to the spatial information using the "label" key. All data is stored in common Python data structures of the \numpy{}, \networkx{}, and \pandas{} libraries.}
    \label{fig:feature_formats}
\end{figure}

Segmentation and tracking information form the basis for extracting spatio-temporal single-cell features. These features include morphological information such as cell location, area, length, perimeter, fluorescence, or cell lifecycle readouts such as age, birth, and division sizes. 
The interpretation of such spatio-temporal features depends highly on the experimental metadata, such as the spatial resolution of the microscope camera, the objective used, and the imaging interval. Therefore, we implemented unit-aware and \textit{modular} cell feature extractors using automated unit computation based on the \pint{} (\url{https://github.com/hgrecco/pint}) library. The segmentation features are extracted for every single cell detection and stored in a unit-aware \pandas{}\cite{team_pandas-devpandas_2024} data frame (\Cref{fig:feature_formats}B). The tracking features are extracted for every single cell cycle and stored in a second \pandas{} table (\Cref{fig:feature_formats}E). Information in both tables is linked using the "label" column, giving a unique number to every cell tracklet.

Using \pandas{} data structures simplifies custom data handling, manipulation, and visualization. The attached physical units are used in downstream computations and facilitate early detection of calculation or calibration errors throughout the analysis pipeline.

\subsection*{Visualization, data analysis and insight generation}

The biological information contained in MLCI time-lapse data is challenging to interpret because it extends across spatial and temporal dimensions, and can be analyzed at the level of single cells, cell colonies, or experiments.
Thus, it is crucial to \textit{customize} analyzes (e.g., growth models or count statistics), and visualizations to generate insights into live-cell imaging data.
Jupyter Notebooks and single-cell feature representation in tabular \pandas{} formats allow the use of the rich visualization tools available in Python, such as Matplotlib~\cite{hunter_matplotlib_2007}, seaborn\cite{waskom_seaborn_2021}, or Plotly (Plotly Technologies Inc., Montreal, Canada,  https://plot.ly). 

In addition to static representations, video rendering and replay capabilities using \moviepy{} (\url{https://github.com/Zulko/moviepy}) is critical for immediate quality control of segmentation and tracking results. This direct visual feedback allows users to adapt or design custom filters, such as filtering out artifacts by selecting a physiologically sensible range of cell sizes.
We also provide the functionality to render cell lineages with attached single-cell information, such as cell size or single-cell growth rates. Combining these videos with charts showing the development of individual cells, colonies, and cell lineages provides comprehensive insights into MLCI data. Interactive exploration of the single-cell data supports the development of new ideas and the uncovering of patterns, raising hypotheses for subsequent quantitative verification through custom data manipulation and visualization~\cite{seiffarth_customizable_2024}.

\subsection*{From pipeline to workflow}

With all these modules having clearly defined interfaces and using standard data formats, we assemble their sequential execution into one Jupyter Notebook. We implement the pipeline, including software dependencies, documentation, and visualizations, and refer to the resulting Notebook as a workflow. \textit{Sharing} this workflow together with imaging data leads to a completely \textit{reproducible} image analysis workflow~\cite{kluyver_jupyter_2016}. These workflows are executed in a web browser, making them \textit{accessible} and easy to \textit{customize}. Conversely, these Notebooks are compatible with high-performance computing setups and are also used at an expert level~\cite{von_chamier_democratising_2021}.

\subsection*{Scaling to quantitative analysis of high-throughput experiments}

Analyzing a single MLCI time-lapse using a workflow gives detailed insights into single-cell development, but it contains limited information on biological heterogeneity and variability. Thus, high-throughput MLCI concurrently observes tens to hundreds of cell populations in a single experiment, offering the potential to quantify heterogeneity and variability across multiple replicates and cell populations.
To apply the same analysis workflow to all these recorded cell populations automatically, we introduce the concept of \textit{scaling} workflows (\Cref{fig:analysis_workflow}D): We parameterize the time-lapse data within the Notebook of a workflow using a unique identifier (e.g. file system path or OMERO ID) and utilize the \papermill{} library (\url{https://github.com/nteract/papermill}) to apply the workflow across multiple time-lapses by automatically updating these parameters. 
This \textit{scaling} procedure is executed in a separate \textit{scaling} workflow, storing the executed workflow for each time-lapse. This documents every analysis step, visualization, and insight for each time-lapse. Within the \textit{scaling} workflow, single-cell data is gathered from all analyzed time-lapses and used to quantify heterogeneity and variability across numerous cell populations.

\section*{Results}

Using the \acia{} library for the composition of the modular image analysis pipeline within workflows, \sixkey{} capabilities are achieved and \SO{} DL tools are available to extract single-cell information at scale. To highlight these new opportunities in single-cell research, we present three application workflows of our workflow collection that analyze publicly available MLCI datasets and extend the analyzes beyond the questions answered in their original publications. These workflows highlight their suitability for analyzing diverse research questions, shedding new light on microbial behavior, and providing new quantitative insights into single-cell development. 

\subsection*{Quantifying microbial population growth}

In biotechnology, the microbial growth rate is the major key performance indicator (KPI) used to evaluate the performance of novel strains under applied cultivation conditions~\cite{kasahara_unveiling_2025,grunberger_beyond_2013}. In MLCI, the biomass (population) growth rate is derived indirectly from changes in cell number or area over time. However, there is no consensus in the literature on which type of growth measure to prefer. Therefore, we present a workflow that computes both cell number and area-based growth rates fully automatically, visualizes them, and highlights the advantages and disadvantages of each measurement approach.

\begin{figure}
    \centering
    \includegraphics[width=0.95\linewidth]{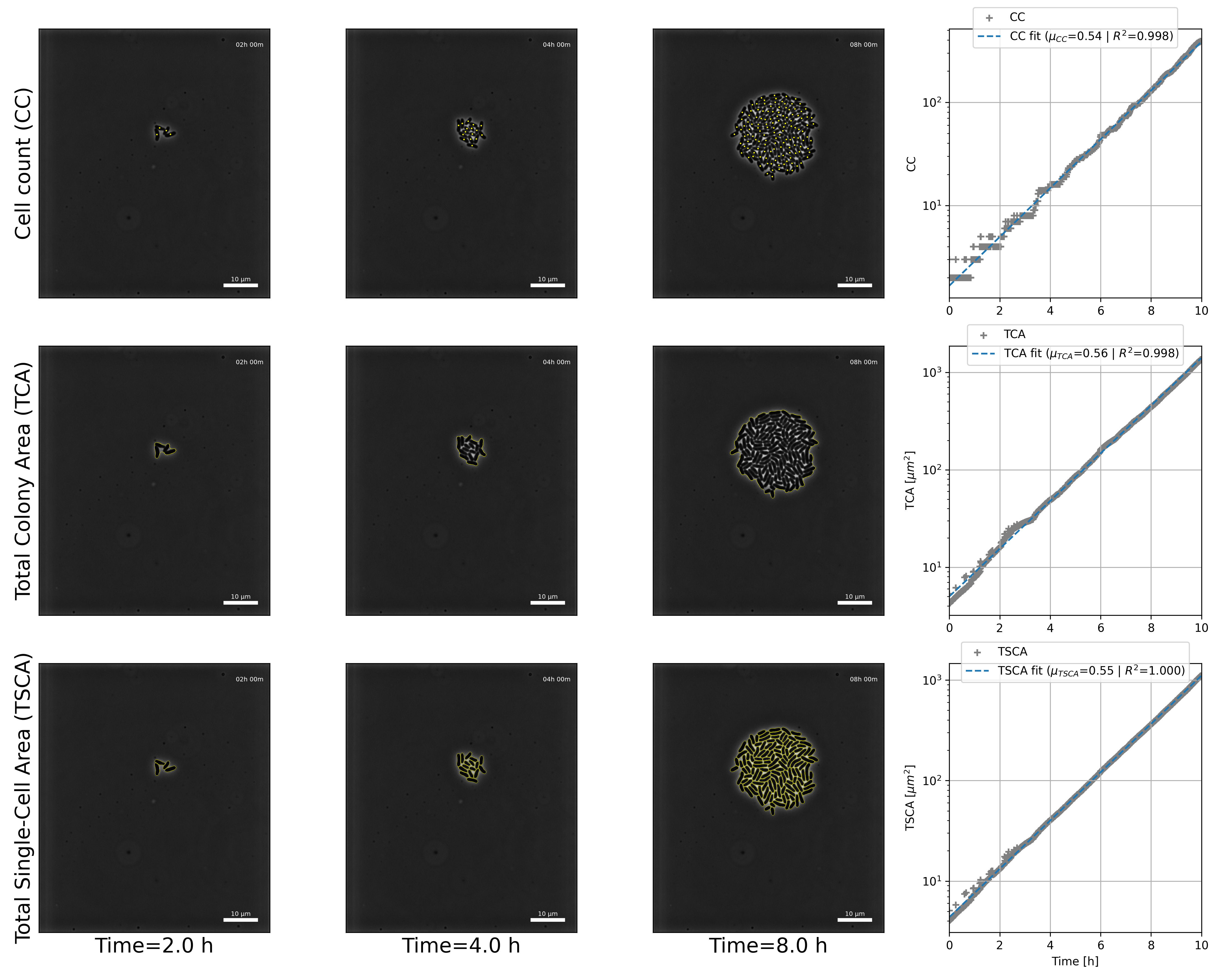}
    \caption{\textbf{Three variants to quantify colony growth rates:} cell count (CC, upper row), total colony area (TCA, middle row), total single-cell area (TSCA, lower row) measured for \CG{} wildtype cultivated in BHI growth medium under constant flow conditions~\citep{Seiffarth2025}. The growth rates are inferred from time-series data (crosses) and fitted (dashed) using a log-linear growth model (right column).}
    \label{fig:microbe_growth}
\end{figure}

In its simplest form, the temporal development of growing cell populations is characterized by counting the cells within a population over time, a task which is sometimes still performed manually~\cite{schmitz_protocol_2024}. 
Using \acia's automated DL-based instance segmentation, both the cell count (CC) and the pixel-precise area of all cells can be extracted automatically from every image. We measure the temporal development of the total single-cell area (TSCA) along with the total colony area (TCA) of the entire colony of cells (\Cref{fig:microbe_growth}). 

We investigate the MLCI time-lapses of cultivating \CG{} under constant environmental conditions~\cite{Seiffarth2025}. Due to the continuous supply of growth medium, we expect to see exponential growth of the cell population. We develop a workflow that performs single-cell segmentation using \texttt{Omnipose}, filters out artifacts by thresholding specific cell sizes, and extracts CC, TCA, and TSCA quantities. We then fit a linear model to these time series in the log-space to determine the three associated exponential growth rates, $\mu_{CC}$, $\mu_{TCA}$, and $\mu_{TSCA}$, respectively.

\Cref{fig:microbe_growth} shows the extraction of the different growth quantities from the time-lapse data, as well as the resulting temporal development and model fits. As expected for this dataset, the cell populations exhibit exponential growth supported by high $R^2$ scores, and all three growth measures show comparable growth rates. However, the cell count quantity (CC) shows discrete step increases, leading to a staircase-like form. These step increases are caused by synchronous cell division, especially within the regime of low cell numbers~\cite{kendall_role_1948,yates_multi-stage_2017,paul_robust_2024,hein_competition_2024}. They become less notable at higher cell numbers. The TCA also exhibits such staircase behavior, albeit much less pronounced than the CC. Interestingly, the TCA quantity includes the area between the cells and, therefore, depends on the packing density of the cell colony. In comparison, the TSCA measures the total single-cell area and shows the smoothest alignment between measured data and log-linear model fit.

Both the TCA and TSCA area-based measures are relatively robust in the event of over-segmentation, as the area of cell instance segmentation is summarized to a population area. In contrast, the CC measure is highly affected by both under- and over-segmentation and should only be used when high-quality segmentation information is available. Our developed workflow enables us to compare all three measures and validate their suitability for the MLCI time-lapse dataset under study.

\subsection*{Characterization of co-cultures using fluorescently labeled strains}

To characterize co-culture development, it is crucial to measure the precise temporal strain-specific composition~\cite{schito_communities_2022,burmeister_microfluidic_2020}. Single-cell measurements in MLCI are ideally suited for this task: labeling bacterial strains with different fluorescence markers allows distinguishing cell strains by their fluorescence signals. 
We develop an application workflow that precisely measures the composition and temporal development of co-cultures, as well as quantifies the growth rates of the individual sub-populations.

We use an MLCI co-culture dataset comprising two \CG{} strains expressing either \mvenus{} or \crimson{} fluorescence proteins (SI 1,2, \url{https://git.nfdi4plants.org/j.seiffarth/bund-et-al_2025}). As the only difference between the two strains is the expressed fluorescence protein (color) and they are cultivated at a constant medium supply, we expect to observe exponential growth with no notable differences in the growth rates of the two strains.

\begin{figure}
    \centering
    \includegraphics[width=0.85\linewidth]{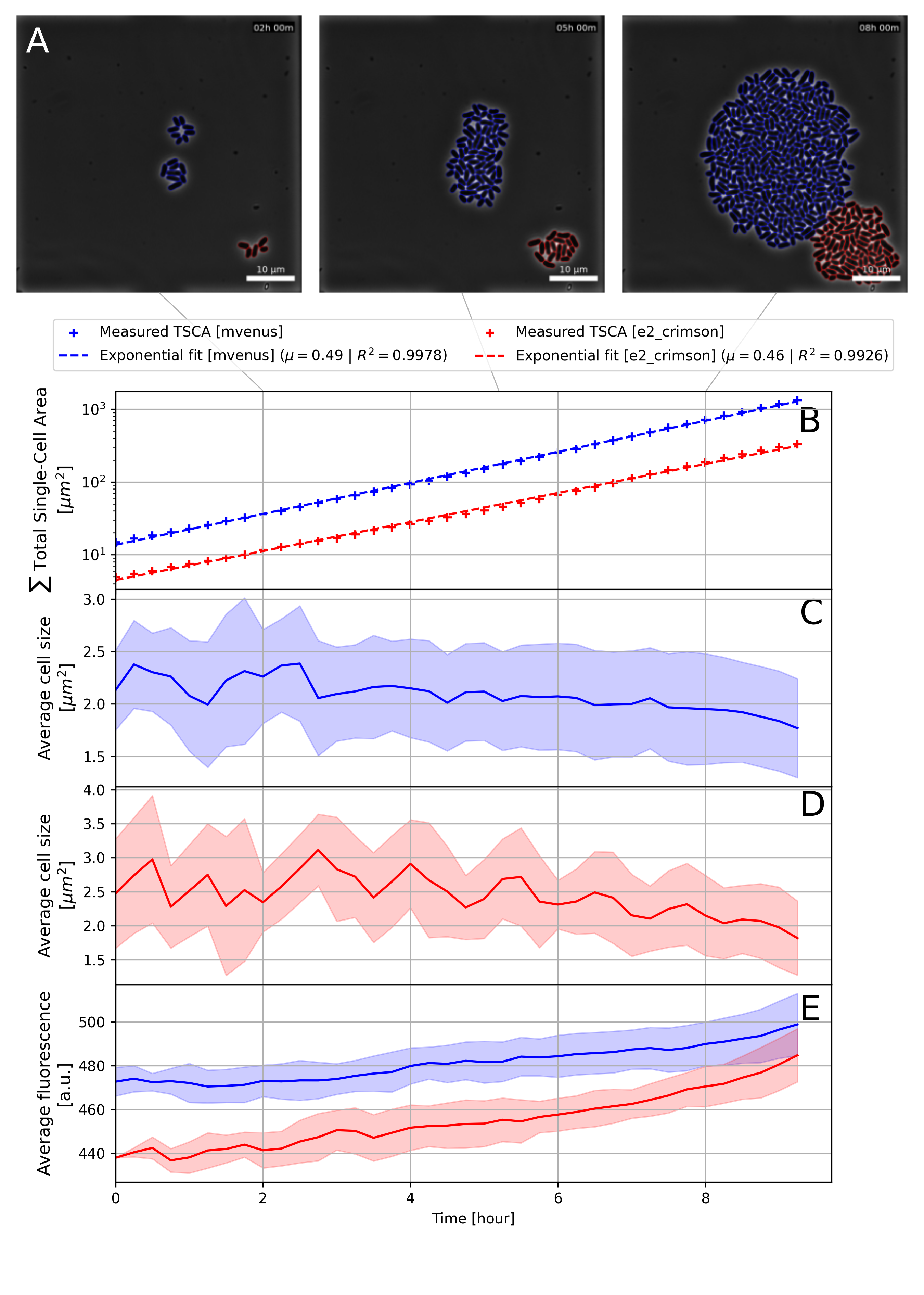}
    \caption{\textbf{Co-culture cultivation with fluorescence-based strain labeling.} Two \CG{} strains with \crimson{} (red) and \mvenus{} (blue) simultaneously cultivated in \texttt{CGXII} medium. (A) Snapshots of the time-lapse with cell outlines colored red and blue depending on their fluorescent label. (B) Temporal development of the measured TSCA of the two sub-populations (crosses) and the fit of an exponential growth model (dashed). (C-D) Average cell area distribution of the sub-populations, with a variance tube (one standard deviation) in light colors. (E) Average fluorescence of the labeled strains (blue/red) with variances (one standard deviation).}
    \label{fig:co_culture_res}
\end{figure}

We develop a workflow that performs segmentation using \texttt{Omnipose} and extracts the mean fluorescence intensity for every cell detection using the fluorescence \texttt{FeatureExtractor} implementation of the \acia{} library. We customize the segmentation, removing non-fluorescent particles (impurities and segmentation artifacts), determining their median cell fluorescence, and apply $k$-means clustering to group the detected cells into two classes (\mvenus{} vs. \crimson{}) using \scipy{}\cite{virtanen_scipy_2020}. \Cref{fig:co_culture_res}A shows excerpts of the time-lapse where the cell outlines are colored based on their fluorescence label. 

As expected, the two \CG{} strains exhibit exponential growth, indicated by the straight regression line in log-space (\Cref{fig:co_culture_res}B). However, the determined TSCA growth rates of the two sub-populations differ slightly (0.46 compared to 0.49~1/h).
More interestingly, the analysis reveals temporal dynamics in both the average cell area and the fluorescence signals (\Cref{fig:co_culture_res}C-E). The average cell area of both sub-populations shows a trend towards an average size of roughly $2~\mu m^2$ for both populations. The smaller colony (red) shows stronger staircase-like fluctuations originating from synchronous cell development and division. These fluctuations are reduced when the number of cells increases.
At the same time, the average fluorescence pixel intensity measured inside the cells increases throughout the experiment (\Cref{fig:co_culture_res}E). We attribute this to the gradual cumulative maturation of fluorescence proteins~\cite{tsien_green_1998}. In summary, our workflow generates insights into the spatio-temporal dynamics of microbial co-cultures.

\subsection*{Quantifying temporal responses of individual cells to environment changes}

For now, we have gained insights into the dynamics of cell populations, but we have not yet investigated and compared the behavior and response of individual cells. To do this, we need to segment and track all cells across the time-lapses.

\begin{figure}
    \centering
    \includegraphics[width=1\linewidth]{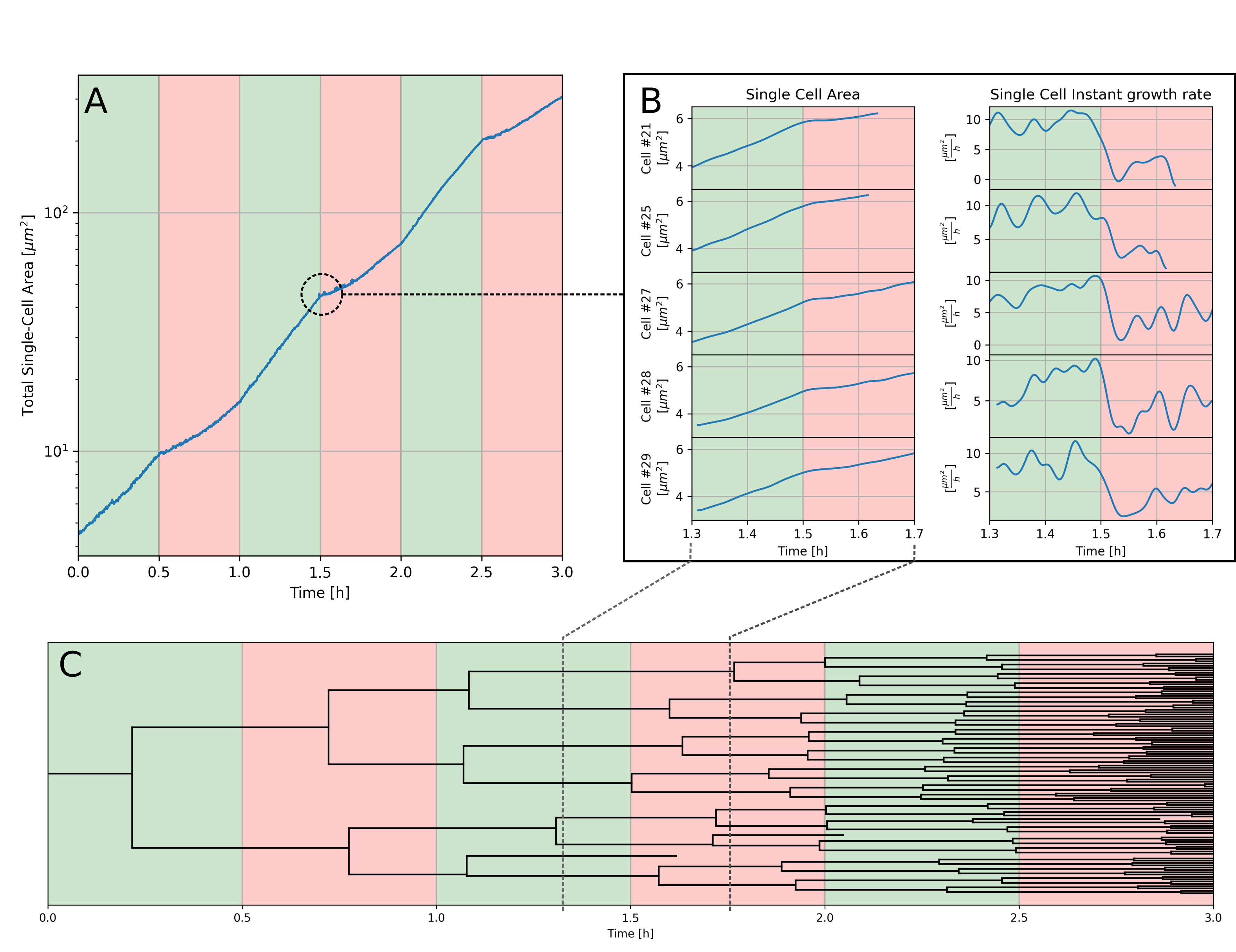}
    \caption{\textbf{Single-cell insights under oxygen switches.} (A) TSCA development of \EC{} as it has been reported by Kasahara et al.\cite{kasahara_unveiling_2025} at alternating aerobic (21\% O\textsubscript{2}, green) and anaerobic (0\% O\textsubscript{2}, red) cultivation conditions. (B) Single-cell area development and instantaneous growth rate (IGR) for five individual cells undergoing the switch from the aerobic (green) to anaerobic (red) cultivation conditions at $t=1.5~h$. (C) Lineage tree generated with automated cell tracking.}
    \label{fig:single_cell_results}
\end{figure}

In our third application workflow, we leverage \acia's capabilities to extract single-cell tracking information to study the impact of oxygen availability on living cells. Anaerobic conditions are ubiquitous in a variety of environments, from the human gut to large-scale bioreactors, and are known to affect growth performance. However, the extent of this impact is difficult to quantify and remains largely unknown. Kasahara et al.~developed a microchip device to perform and collect MLCI data of \EC{} under precisely controlled and fast (approximately 10s) oxygen switches between $0\%$ and $21\%$ oxygen~\cite{kasahara_unveiling_2025}. They observed that the \EC{} populations respond within minutes to the removal of oxygen by reducing their TSCA growth rate (\Cref{fig:single_cell_results}A). Thus, we develop a workflow that extends this analysis to the single-cell level and confirm that this effect is also present for the single cells. Moreover, we investigate the consistency of the cells' responses in terms of timing and adaptation to these alternating oxygen conditions.

To this end, we develop a workflow that uses \textit{Omnipose} segmentation, followed by cell tracking using the \textit{trackastra} cell tracker. To remove artifacts, we first correct for over-segmentation and only investigate cells observed throughout their entire life cycle (i.e., from birth to division). We then use \acia's \texttt{FeatureExtrator} to extract and quantify the single-cell area. As a reference, \Cref{fig:single_cell_results}A shows the temporal TSCA development, reproducing the results reported by Kasahara et al.\cite{kasahara_unveiling_2025}. Phases of 21\% oxygen are colored green, and phases of 0\% oxygen are colored red. In their study, Kasahara et al.~observed a substantial decrease in TSCA growth rate upon entering the anaerobic phase (e.g. at $t=1.5~h$). However, the data in terms of the TSCA-based growth rate obscures differences in individual cell responses within the same population, which may differ substantially in terms of their strength and timing.

Tracking individual cells gives insights into their temporal reactions to changes in oxygen levels. \Cref{fig:single_cell_results}C shows the lineage tree reconstructed by the cell tracking. The lineage shows how a single cell within the population (left)  develops into more and more cells due to cell divisions, as indicated by the branches in the lineage. Based on this lineage, we select cells that have been observed before and after entry into the anaerobic phase at $t=1.5~h$. 
\Cref{fig:single_cell_results}B shows the development of five exemplary cells undergoing the switch from aerobic to anaerobic conditions ($t=1.5~h$). We measure their individual area and their instantaneous growth rate (IGR), i.e., the time derivatives of the single-cell area development (SI 3). Interestingly, the IGR shows a strong and rapid decrease shortly after the switch event and reaches a local minimum within minutes after the oxygen change. Notably, our workflow is able to show that this rapid response to the change in oxygen availability is consistent across all five cells and thus highlights the new opportunities in uncovering single-cell dynamics.

\subsection*{Scaling up to quantitative insights in high-throughput MLCI}

The application workflows presented so far extract detailed single-cell information from a single MLCI time-lapse. To assess the variability of single-cell behavior and the significance of the insights gained, we exploit the high-throughput capabilities of MLCI by recording time-lapse videos of multiple cell populations and analyzing them using \acia's workflow scaling functionality. 
Here, the developed workflows are automatically rolled out to all replicates of an experiment, and the quantitative results are summarized and visualized.

\begin{figure}
    \centering
    \includegraphics[width=0.95\linewidth]{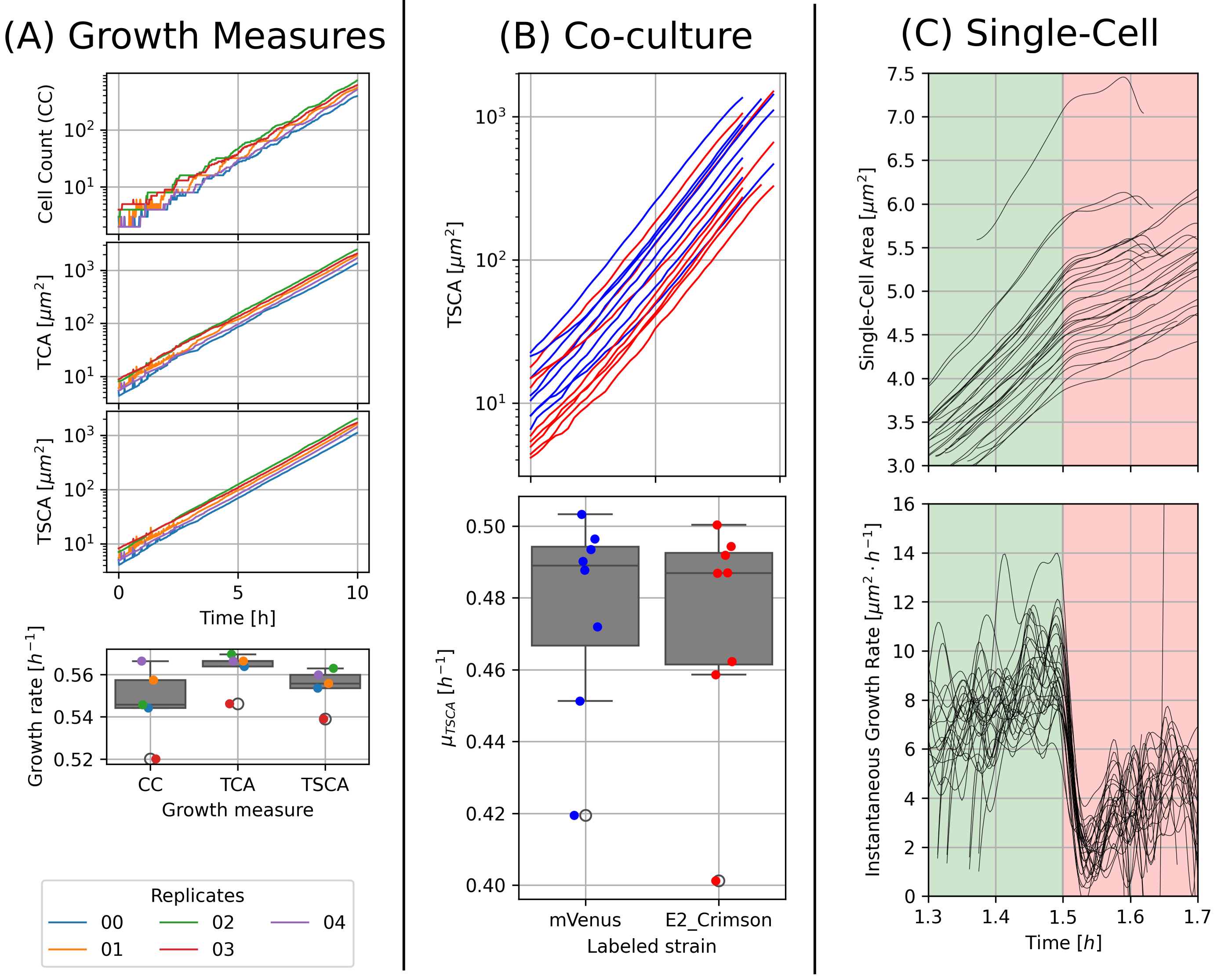}
    \caption{\textbf{Scaling MLCI analysis to multiple replicates.} (A) shows CC, TCA, and TCA development (top to bottom) for five \CG{} replicates. The growth rate distribution for all three measures is shown at the bottom. Measurements for the replicates are shown in different colors. (B) shows the \CG{} co-culture TSCA development (top) and exponential growth rate quantification based on TSCA (bottom) for eight replicates for \mvenus{} (blue) and \crimson{} (red) labeled strains. (C) shows the single-cell area development (top) and instantaneous growth rates (bottom) of \EC{} in aerobic (green) and anaerobic phase (red) for five replicates and 30 individual cells.}
    \label{fig:scaling_results}
\end{figure}

In our first workflow, we have compared various types of growth rate measurements. By analyzing multiple time-lapses, we now investigate the variability in these growth rate measurements. \Cref{fig:scaling_results}A shows the quantitative analysis of population growth measures with five replicates for CC, TCA, and TSCA measures, including growth rates derived by regressing the measurements using a log-linear model (top to bottom). The temporal development of all three quantities is well-described by the log-linear model, and all three growth measures give growth rates in the range of $0.52~h^{-1}$ to $0.57~h^{-1}$. Moreover, the analysis confirmed differences between count-based and area-based population quantification for low cell counts: the two area-based measurements (TCA, TSCA) show almost no fluctuations compared to the CC measure, which shows the familiar staircase-like increase due to synchronous cell divisions. Furthermore, replicate "03" consistently shows the lowest growth rate across all three measurements and might qualify as an outlier. However, more replicates are necessary to make this assessment. In summary, scaling the growth measurement workflow to all five time-lapses allows quantifying the variation in measured growth rates and identifying potential outliers.

In our second workflow, we have demonstrated the ability to precisely measure the growth rates of co-culture populations using \mvenus- and \crimson-labeled \CG{} strains as an example. This raises the question of whether there is a systematic difference in their growth rates caused by the two fluorescence reporters. To quantify a potential deviation in the growth rate measurements, we need to analyze multiple cultivations. 

\Cref{fig:scaling_results}B shows the quantitative growth rate analysis of the fluorescence-labeled strains with eight replicates using the TSCA measure (top), as well as the inferred growth rates (bottom).
Our analysis yields average TSCA growth rates of $0.477~h^{-1} \pm 0.026$ for \mvenus{} and $0.473~h^{-1} \pm 0.030$ for the \crimson{} strains, respectively. Despite the slightly higher average growth rate of the \mvenus{} strains, quantification of the variability in measured growth rates shows that the growth rate of both strains is not systematically different. We conclude that encoding the two fluorescence proteins does not lead to a measurable difference in growth rates.

For the last workflow, we have shown that individual cells within a cell populations react consistently within minutes to changes in oxygen availability by reducing their IGR. To ensure that this effect occurs consistently across different cell populations, we roll out the analysis workflow to all five replicates in the dataset. 
We analyze the response of \EC{} upon switching from the aerobic phase to the anaerobic phase in the experiment ($t=1.5~h$). Employing single-cell tracking, \Cref{fig:scaling_results}C shows the development of the single-cell area (top) and the IGR at the single-cell level (bottom) for 30 cells. 
As expected, cell growth is faster during the aerobic phase (green) than in the anaerobic phase (red). Moreover, all individual cells exhibit variations in their IGR (bottom) before entering the anaerobic phase. However, upon entering the anaerobic phase, we observe an immediate steep drop in IGR for all cells across all replicates. After this initial steep drop, the cells again show variations in their IGR. Despite large fluctuations in single-cell growth rates (i.e., IGR) in constant oxygen regimes, all cells respond with a consistent and rapid decrease in IGR to oxygen removal from their environment.

All three examples show that the capability to easily \textit{scale} the developed workflows is crucial to unlocking quantitative analyzes of living cells and builds the foundation for more robust and reproducible insights.

\section*{Conclusion \& Outlook}

The analysis of high-throughput time-lapse data poses high requirements on the image analysis pipeline, including \textit{customizability}, robustness, and \textit{scalability} of analysis workflows to extract insights from time-lapse data reliably. 
With the introduced \aw{} platform, we combine \SO{} image processing and data extraction in modular image analysis pipelines and compose these components 
in Jupyter Notebook-based workflows that provide a modular analysis pipeline with extensive customization, are easy to use, fully \textit{reproducible}, and \textit{scale} to high-throughput screenings. We demonstrate the capabilities of the \aw{} platform with three workflow applications generating new insights into the behavior of single cells and their variability.

For now, \acia{} is primarily designed to process 2D+time time-lapses, however, we plan to implement modules for 3D segmentation and tracking methods in the future to extend the analysis capabilities to 3D image sequences. Moreover, our workflow applications focused on the analysis of bacterial cells. However, \acia{} is not limited to bacterial cells, and workflows for other cell morphologies and imaging modalities, for example, from the cell tracking challenge, are available in our workflow collection.

Customizing workflows requires users to write Python code. While writing code may come with a steep learning curve for a domain scientist, large language models are effective in writing short Python snippets, e.g., for custom visualizations or data handling~\cite{royer_omega_2024}. Thus, we believe that the \textit{accessibility} of Jupyter Notebooks, executed online in the web browser, outweighs the challenges of learning to write short Python code snippets. Moreover, the code precisely documents every step in the pipeline and allows others to understand and \textit{share} existing code snippets.

Publishing the \aw{} platform with its collection of application workflows, we have lowered the entry barrier of MLCI analysis while showing \textit{customizability} and \textit{reproducibility} for high-throughput analyzes. Thus, \aw{} represents an important step toward democratizing image analysis workflows in live-cell imaging, unlocking the potential of single-cell insights for life scientists and data scientists alike.

\subsubsection*{Data and code availability}
This paper analyzes existing, publicly available time-lapse data.
The "Tracking One in A Million" dataset~\cite{Seiffarth2025} is available at zenodo: \url{https://doi.org/10.5281/zenodo.7260136}, the co-culture dataset is available at PLANTdataHUB\cite{weil_plantdatahub_2023}: \url{https://git.nfdi4plants.org/j.seiffarth/bund-et-al_2025}, and the oxygen switching dataset is available at zenodo: \url{https://doi.org/10.5281/zenodo.13982746}. 
All original code for the \acia{} Python library is available at \url{https://github.com/JuBiotech/acia-core} and the workflow collection available at \url{https://github.com/JuBiotech/acia-workflows} in form of Jupyter Notebooks. Both code repositories are published under MIT license.

\section*{Supplemental information}
\label{sec:SI}
\begin{description}
  \item S.1 Strain, plasmid and oligonucleotide details.
  \item S.2 Microfluidic cultivation details.
  \item S.3 Single-cell analysis details.
\end{description}

\section*{Acknowledgments}
J.S. was supported by the President's Initiative and Networking Funds of the Helmholtz Association of German Research Centres [EMSIG ZT-I-PF-04-44] and received  funding from the Helmholtz Association of German Research Centres within the Helmholtz School for Data Science in Life, Earth, and Energy (HDS-LEE).
R.P. received  funding from the Helmholtz Association of German Research Centres within the Helmholtz School for Data Science in Life, Earth, and Energy (HDS-LEE).
K.K., M.Bu., B.L., and L.W. were funded by the Deutsche Forschungsgemeinschaft (DFG, German Research Foundation) – SFB1535 - Project ID 458090666. 


\section*{Declaration of interests}
The authors declare no competing interests.

\bibliography{osw}

\begin{thebibliography}{64}
\providecommand{\natexlab}[1]{#1}
\providecommand{\url}[1]{\texttt{#1}}
\providecommand{\href}[2]{#2}
\providecommand{\path}[1]{#1}
\providecommand{\DOIprefix}{doi:}
\providecommand{\ArXivprefix}{arXiv:}
\providecommand{\URLprefix}{URL: }
\providecommand{\Pubmedprefix}{pmid:}
\providecommand{\doi}[1]{\href{http://dx.doi.org/#1}{\path{#1}}}
\providecommand{\Pubmed}[1]{\href{pmid:#1}{\path{#1}}}
\providecommand{\BIBand}{and}
\providecommand{\bibinfo}[2]{#2}
\ifx\xfnm\undefined \def\xfnm[#1]{\unskip,\space#1}\fi
\makeatletter\def\@biblabel#1{#1.}\makeatother
\bibitem[{Alieva et~al.(2023)Alieva, Wezenaar, Wehrens and Rios}]{alieva_bridging_2023}
\bibinfo{author}{Alieva, M.}, \bibinfo{author}{Wezenaar, A. K.~L.}, \bibinfo{author}{Wehrens, E.~J.}, and \bibinfo{author}{Rios, A.~C.} (\bibinfo{year}{2023}). \bibinfo{title}{Bridging live-cell imaging and next-generation cancer treatment}.
\newblock \bibinfo{journal}{Nature Reviews Cancer} \emph{\bibinfo{volume}{23}}, \bibinfo{pages}{731--745}. \DOIprefix\doi{10.1038/s41568-023-00610-5}.
\bibitem[{Shirasaki et~al.(2014)Shirasaki, Yamagishi, Suzuki, Izawa, Nakahara, Mizuno, Shoji, Heike, Harada, Nishikomori and Ohara}]{shirasaki_real-time_2014}
\bibinfo{author}{Shirasaki, Y.}, \bibinfo{author}{Yamagishi, M.}, \bibinfo{author}{Suzuki, N.}, \bibinfo{author}{Izawa, K.}, \bibinfo{author}{Nakahara, A.}, \bibinfo{author}{Mizuno, J.}, \bibinfo{author}{Shoji, S.}, \bibinfo{author}{Heike, T.}, \bibinfo{author}{Harada, Y.}, \bibinfo{author}{Nishikomori, R.}, and \bibinfo{author}{Ohara, O.} (\bibinfo{year}{2014}). \bibinfo{title}{Real-time single-cell imaging of protein secretion}.
\newblock \bibinfo{journal}{Scientific Reports} \emph{\bibinfo{volume}{4}}, \bibinfo{pages}{4736}. \DOIprefix\doi{10.1038/srep04736}.
\bibitem[{Raphael et~al.(2013)Raphael, Christodoulides, Delehanty, Long and Byers}]{raphael_quantitative_2013}
\bibinfo{author}{Raphael, M.~P.}, \bibinfo{author}{Christodoulides, J.~A.}, \bibinfo{author}{Delehanty, J.~B.}, \bibinfo{author}{Long, J.~P.}, and \bibinfo{author}{Byers, J.~M.} (\bibinfo{year}{2013}). \bibinfo{title}{Quantitative imaging of protein secretions from single cells in real time}.
\newblock \bibinfo{journal}{Biophysical Journal} \emph{\bibinfo{volume}{105}}, \bibinfo{pages}{602--608}. \DOIprefix\doi{10.1016/j.bpj.2013.06.022}.
\bibitem[{Weissmann and Brandt(2008)}]{weissmann_mechanisms_2008}
\bibinfo{author}{Weissmann, C.}, and \bibinfo{author}{Brandt, R.} (\bibinfo{year}{2008}). \bibinfo{title}{Mechanisms of neurodegenerative diseases: {Insights} from live cell imaging}.
\newblock \bibinfo{journal}{Journal of Neuroscience Research} \emph{\bibinfo{volume}{86}}, \bibinfo{pages}{504--511}. \DOIprefix\doi{10.1002/jnr.21448}.
\bibitem[{Campbell and Hope(2008)}]{campbell_live_2008}
\bibinfo{author}{Campbell, E.~M.}, and \bibinfo{author}{Hope, T.~J.} (\bibinfo{year}{2008}). \bibinfo{title}{Live cell imaging of the {HIV}-1 life cycle}.
\newblock \bibinfo{journal}{Trends in Microbiology} \emph{\bibinfo{volume}{16}}, \bibinfo{pages}{580--587}. \DOIprefix\doi{10.1016/j.tim.2008.09.006}.
\bibitem[{Preedy et~al.(2024)Preedy, White and Tergaonkar}]{preedy_cellular_2024}
\bibinfo{author}{Preedy, M.~K.}, \bibinfo{author}{White, M. R.~H.}, and \bibinfo{author}{Tergaonkar, V.} (\bibinfo{year}{2024}). \bibinfo{title}{Cellular heterogeneity in {TNF}/{TNFR1} signalling: {L}ive cell imaging of cell fate decisions in single cells}.
\newblock \bibinfo{journal}{Cell Death \& Disease} \emph{\bibinfo{volume}{15}}, \bibinfo{pages}{1--12}. \DOIprefix\doi{10.1038/s41419-024-06559-z}.
\bibitem[{Huang et~al.(2024)Huang, Zhou, Liu, Tang and Xin}]{huang_exploring_2024}
\bibinfo{author}{Huang, Y.}, \bibinfo{author}{Zhou, Z.}, \bibinfo{author}{Liu, T.}, \bibinfo{author}{Tang, S.}, and \bibinfo{author}{Xin, X.} (\bibinfo{year}{2024}). \bibinfo{title}{Exploring heterogeneous cell population dynamics in different microenvironments by novel analytical strategy based on images}.
\newblock \bibinfo{journal}{npj Systems Biology and Applications} \emph{\bibinfo{volume}{10}}, \bibinfo{pages}{1--10}. \DOIprefix\doi{10.1038/s41540-024-00459-w}.
\bibitem[{Drescher et~al.(2014)Drescher, Nadell, Stone, Wingreen and Bassler}]{drescher_solutions_2014}
\bibinfo{author}{Drescher, K.}, \bibinfo{author}{Nadell, C.~D.}, \bibinfo{author}{Stone, H.~A.}, \bibinfo{author}{Wingreen, N.~S.}, and \bibinfo{author}{Bassler, B.~L.} (\bibinfo{year}{2014}). \bibinfo{title}{Solutions to the public goods dilemma in bacterial biofilms}.
\newblock \bibinfo{journal}{Current Biology} \emph{\bibinfo{volume}{24}}, \bibinfo{pages}{50--55}. \DOIprefix\doi{10.1016/j.cub.2013.10.030}.
\bibitem[{Hartmann et~al.(2019)Hartmann, Singh, Pearce, Mok, Song, Díaz-Pascual, Dunkel and Drescher}]{hartmann_emergence_2019}
\bibinfo{author}{Hartmann, R.}, \bibinfo{author}{Singh, P.~K.}, \bibinfo{author}{Pearce, P.}, \bibinfo{author}{Mok, R.}, \bibinfo{author}{Song, B.}, \bibinfo{author}{Díaz-Pascual, F.}, \bibinfo{author}{Dunkel, J.}, and \bibinfo{author}{Drescher, K.} (\bibinfo{year}{2019}). \bibinfo{title}{Emergence of three-dimensional order and structure in growing biofilms}.
\newblock \bibinfo{journal}{Nature Physics} \emph{\bibinfo{volume}{15}}, \bibinfo{pages}{251--256}. \DOIprefix\doi{10.1038/s41567-018-0356-9}.
\bibitem[{Cornaglia et~al.(2017)Cornaglia, Lehnert and M.~Gijs}]{cornaglia_microfluidic_2017}
\bibinfo{author}{Cornaglia, M.}, \bibinfo{author}{Lehnert, T.}, and \bibinfo{author}{M.~Gijs, M.~A.} (\bibinfo{year}{2017}). \bibinfo{title}{Microfluidic systems for high-throughput and high-content screening using the nematode \textit{{C}aenorhabditis elegans}}.
\newblock \bibinfo{journal}{Lab on a Chip} \emph{\bibinfo{volume}{17}}, \bibinfo{pages}{3736--3759}. \DOIprefix\doi{10.1039/C7LC00509A}.
\bibitem[{Ortseifen et~al.(2020)Ortseifen, Viefhues, Wobbe and Gr{\"{u}}nberger}]{ortseifen_microfluidics_2020}
\bibinfo{author}{Ortseifen, V.}, \bibinfo{author}{Viefhues, M.}, \bibinfo{author}{Wobbe, L.}, and \bibinfo{author}{Gr{\"{u}}nberger, A.} (\bibinfo{year}{2020}). \bibinfo{title}{Microfluidics for biotechnology: {B}ridging gaps to foster microfluidic applications}.
\newblock \bibinfo{journal}{Frontiers in Bioengineering and Biotechnology} \emph{\bibinfo{volume}{8}}. \DOIprefix\doi{10.3389/fbioe.2020.589074}.
\bibitem[{Weibel et~al.(2007)Weibel, DiLuzio and Whitesides}]{weibel_microfabrication_2007}
\bibinfo{author}{Weibel, D.~B.}, \bibinfo{author}{DiLuzio, W.~R.}, and \bibinfo{author}{Whitesides, G.~M.} (\bibinfo{year}{2007}). \bibinfo{title}{Microfabrication meets microbiology}.
\newblock \bibinfo{journal}{Nature Reviews Microbiology} \emph{\bibinfo{volume}{5}}, \bibinfo{pages}{209--218}. \DOIprefix\doi{10.1038/nrmicro1616}.
\bibitem[{Witting et~al.(2025)Witting, Seiffarth, Stute, Schulze, Hofer, Nöh, Eisenhut, Weber, Lieres and Kohlheyer}]{witting_microfluidic_2025}
\bibinfo{author}{Witting, L.}, \bibinfo{author}{Seiffarth, J.}, \bibinfo{author}{Stute, B.}, \bibinfo{author}{Schulze, T.}, \bibinfo{author}{Hofer, J.~M.}, \bibinfo{author}{Nöh, K.}, \bibinfo{author}{Eisenhut, M.}, \bibinfo{author}{Weber, A. P.~M.}, \bibinfo{author}{Lieres, E.~v.}, and \bibinfo{author}{Kohlheyer, D.} (\bibinfo{year}{2025}). \bibinfo{title}{A microfluidic system for the cultivation of cyanobacteria with precise light intensity and {CO}\textsubscript{2} control: {E}nabling growth data acquisition at single-cell resolution}.
\newblock \bibinfo{journal}{Lab on a Chip} \emph{\bibinfo{volume}{25}}, \bibinfo{pages}{319--329}. \DOIprefix\doi{10.1039/D4LC00567H}.
\bibitem[{Kasahara et~al.(2025)Kasahara, Seiffarth, Stute, von Lieres, Drepper, N{\"{o}}h and Kohlheyer}]{kasahara_unveiling_2025}
\bibinfo{author}{Kasahara, K.}, \bibinfo{author}{Seiffarth, J.}, \bibinfo{author}{Stute, B.}, \bibinfo{author}{von Lieres, E.}, \bibinfo{author}{Drepper, T.}, \bibinfo{author}{N{\"{o}}h, K.}, and \bibinfo{author}{Kohlheyer, D.} (\bibinfo{year}{2025}). \bibinfo{title}{{Unveiling microbial single-cell growth dynamics under rapid periodic oxygen oscillations}}.
\newblock \bibinfo{journal}{Lab on a Chip} \emph{\bibinfo{volume}{25}}, \bibinfo{pages}{2234--2246}. \DOIprefix\doi{10.1039/D5LC00065C}.
\bibitem[{Täuber et~al.(2020)Täuber, Golze, Ho, Lieres and Grünberger}]{tauber_dmscc_2020}
\bibinfo{author}{Täuber, S.}, \bibinfo{author}{Golze, C.}, \bibinfo{author}{Ho, P.}, \bibinfo{author}{Lieres, E.~v.}, and \bibinfo{author}{Grünberger, A.} (\bibinfo{year}{2020}). \bibinfo{title}{{dMSCC}: {A} microfluidic platform for microbial single-cell cultivation of \textit{{C}orynebacterium glutamicum} under dynamic environmental medium conditions}.
\newblock \bibinfo{journal}{Lab on a Chip} \emph{\bibinfo{volume}{20}}, \bibinfo{pages}{4442--4455}. \DOIprefix\doi{10.1039/D0LC00711K}.
\bibitem[{Schito et~al.(2022)Schito, Zuchowski, Bergen, Strohmeier, Wollenhaupt, Menke, Seiffarth, Nöh, Kohlheyer, Bott, Wiechert, Baumgart and Noack}]{schito_communities_2022}
\bibinfo{author}{Schito, S.}, \bibinfo{author}{Zuchowski, R.}, \bibinfo{author}{Bergen, D.}, \bibinfo{author}{Strohmeier, D.}, \bibinfo{author}{Wollenhaupt, B.}, \bibinfo{author}{Menke, P.}, \bibinfo{author}{Seiffarth, J.}, \bibinfo{author}{Nöh, K.}, \bibinfo{author}{Kohlheyer, D.}, \bibinfo{author}{Bott, M.}, \bibinfo{author}{Wiechert, W.}, \bibinfo{author}{Baumgart, M.}, and \bibinfo{author}{Noack, S.} (\bibinfo{year}{2022}). \bibinfo{title}{Communities of {Niche}-optimized {Strains} ({CoNoS}) – {Design} and creation of stable, genome-reduced co-cultures}.
\newblock \bibinfo{journal}{Metabolic Engineering} \emph{\bibinfo{volume}{73}}, \bibinfo{pages}{91--103}. \DOIprefix\doi{10.1016/j.ymben.2022.06.004}.
\bibitem[{Burmeister and Grünberger(2020)}]{burmeister_microfluidic_2020}
\bibinfo{author}{Burmeister, A.}, and \bibinfo{author}{Grünberger, A.} (\bibinfo{year}{2020}). \bibinfo{title}{Microfluidic cultivation and analysis tools for interaction studies of microbial co-cultures}.
\newblock \bibinfo{journal}{Current Opinion in Biotechnology} \emph{\bibinfo{volume}{62}}, \bibinfo{pages}{106--115}. \DOIprefix\doi{10.1016/j.copbio.2019.09.001}.
\bibitem[{Chung et~al.(2024)Chung, Kar, Kamkaew, Amir and Aldridge}]{chung_single-cell_2024}
\bibinfo{author}{Chung, E.~S.}, \bibinfo{author}{Kar, P.}, \bibinfo{author}{Kamkaew, M.}, \bibinfo{author}{Amir, A.}, and \bibinfo{author}{Aldridge, B.~B.} (\bibinfo{year}{2024}). \bibinfo{title}{Single-cell imaging of the \textit{{M}ycobacterium tuberculosis} cell cycle reveals linear and heterogenous growth}.
\newblock \bibinfo{journal}{Nature Microbiology} \emph{\bibinfo{volume}{9}}, \bibinfo{pages}{3332--3344}. \DOIprefix\doi{10.1038/s41564-024-01846-z}.
\bibitem[{Jeckel and Drescher(2021)}]{jeckel_advances_2021}
\bibinfo{author}{Jeckel, H.}, and \bibinfo{author}{Drescher, K.} (\bibinfo{year}{2021}). \bibinfo{title}{Advances and opportunities in image analysis of bacterial cells and communities}.
\newblock \bibinfo{journal}{FEMS Microbiology Reviews} \emph{\bibinfo{volume}{45}}. \DOIprefix\doi{10.1093/femsre/fuaa062}.
\bibitem[{Yazdi and Khotanlou(2024)}]{yazdi_survey_2024}
\bibinfo{author}{Yazdi, R.}, and \bibinfo{author}{Khotanlou, H.} (\bibinfo{year}{2024}). \bibinfo{title}{A survey on automated cell tracking: {C}hallenges and solutions}.
\newblock \bibinfo{journal}{Multimedia Tools and Applications} \emph{\bibinfo{volume}{83}}, \bibinfo{pages}{81511--81547}. \DOIprefix\doi{10.1007/s11042-024-18697-9}.
\bibitem[{Cutler et~al.(2022)Cutler, Stringer, Lo, Rappez, Stroustrup, Brook~Peterson, Wiggins and Mougous}]{cutler_omnipose_2022}
\bibinfo{author}{Cutler, K.~J.}, \bibinfo{author}{Stringer, C.}, \bibinfo{author}{Lo, T.~W.}, \bibinfo{author}{Rappez, L.}, \bibinfo{author}{Stroustrup, N.}, \bibinfo{author}{Brook~Peterson, S.}, \bibinfo{author}{Wiggins, P.~A.}, and \bibinfo{author}{Mougous, J.~D.} (\bibinfo{year}{2022}). \bibinfo{title}{Omnipose: {A} high-precision morphology-independent solution for bacterial cell segmentation}.
\newblock \bibinfo{journal}{Nature Methods} \emph{\bibinfo{volume}{19}}, \bibinfo{pages}{1438--1448}. \DOIprefix\doi{10.1038/s41592-022-01639-4}.
\bibitem[{Stringer et~al.(2021)Stringer, Wang, Michaelos and Pachitariu}]{stringer_cellpose_2021}
\bibinfo{author}{Stringer, C.}, \bibinfo{author}{Wang, T.}, \bibinfo{author}{Michaelos, M.}, and \bibinfo{author}{Pachitariu, M.} (\bibinfo{year}{2021}). \bibinfo{title}{{C}ellpose: {A} generalist algorithm for cellular segmentation}.
\newblock \bibinfo{journal}{Nature Methods} \emph{\bibinfo{volume}{18}}, \bibinfo{pages}{100--106}. \DOIprefix\doi{10.1038/s41592-020-01018-x}.
\bibitem[{Upschulte et~al.(2023)Upschulte, Harmeling, Amunts and Dickscheid}]{upschulte_uncertainty-aware_2023}
\bibinfo{author}{Upschulte, E.}, \bibinfo{author}{Harmeling, S.}, \bibinfo{author}{Amunts, K.}, and \bibinfo{author}{Dickscheid, T.} (\bibinfo{year}{2023}).
\newblock \bibinfo{title}{Uncertainty-aware contour proposal networks for cell segmentation in multi-modality high-resolution microscopy images}.
\newblock In: \bibinfo{booktitle}{Proceedings of {The} {Cell} {Segmentation} {Challenge} in {Multi}-modality {High}-{Resolution} {Microscopy} {Images}}. \bibinfo{publisher}{PMLR} ( \bibinfo{pages}{1--12}).
\newblock \URLprefix \url{https://proceedings.mlr.press/v212/upschulte23a.html}.
\bibitem[{Seiffarth et~al.(2025)Seiffarth, Bl{\"{o}}baum, Paul, Friederich, Sitcheu, Mikut, Scharr, Gr{\"{u}}nberger and N{\"{o}}h}]{Seiffarth2025}
\bibinfo{author}{Seiffarth, J.}, \bibinfo{author}{Bl{\"{o}}baum, L.}, \bibinfo{author}{Paul, R.~D.}, \bibinfo{author}{Friederich, N.}, \bibinfo{author}{Sitcheu, A. J.~Y.}, \bibinfo{author}{Mikut, R.}, \bibinfo{author}{Scharr, H.}, \bibinfo{author}{Gr{\"{u}}nberger, A.}, and \bibinfo{author}{N{\"{o}}h, K.} (\bibinfo{year}{2025}).
\newblock \bibinfo{title}{{Tracking One-in-a-Million: Large-Scale Benchmark for Microbial Single-Cell Tracking with Experiment-Aware Robustness Metrics}}.
\newblock In: \bibinfo{editor}{{Del Bue}, A.}, \bibinfo{editor}{Canton, C.}, \bibinfo{editor}{Pont-Tuset, J.}, and \bibinfo{editor}{Tommasi, T.}, eds. \bibinfo{booktitle}{Computer Vision – ECCV 2024 Workshops. ECCV 2024. Lecture Notes in Computer Science, vol 15638}. \bibinfo{publisher}{Springer, Cham} ( \bibinfo{pages}{318--334}).
\newblock \DOIprefix\doi{10.1007/978-3-031-91721-9_20}.
\bibitem[{Scherr et~al.(2022)Scherr, Seiffarth, Wollenhaupt, Neumann, Schilling, Kohlheyer, Scharr, Nöh and Mikut}]{scherr_microbeseg_2022}
\bibinfo{author}{Scherr, T.}, \bibinfo{author}{Seiffarth, J.}, \bibinfo{author}{Wollenhaupt, B.}, \bibinfo{author}{Neumann, O.}, \bibinfo{author}{Schilling, M.~P.}, \bibinfo{author}{Kohlheyer, D.}, \bibinfo{author}{Scharr, H.}, \bibinfo{author}{Nöh, K.}, and \bibinfo{author}{Mikut, R.} (\bibinfo{year}{2022}). \bibinfo{title}{{microbeSEG}: {A} deep learning software tool with {OMERO} data management for efficient and accurate cell segmentation}.
\newblock \bibinfo{journal}{PLOS ONE} \emph{\bibinfo{volume}{17}}, \bibinfo{pages}{e0277601}. \DOIprefix\doi{10.1371/journal.pone.0277601}.
\bibitem[{Gallusser and Weigert(2024)}]{gallusser_trackastra_2024}
\bibinfo{author}{Gallusser, B.}, and \bibinfo{author}{Weigert, M.} (\bibinfo{year}{2024}).
\newblock \bibinfo{title}{Trackastra: {Transformer}-based cell tracking for live-cell microscopy}.
\newblock \bibinfo{publisher}{arXiv [cs]}.
\newblock \DOIprefix\doi{10.48550/arXiv.2405.15700}.
\bibitem[{Jaqaman et~al.(2008)Jaqaman, Loerke, Mettlen, Kuwata, Grinstein, Schmid and Danuser}]{jaqaman_robust_2008}
\bibinfo{author}{Jaqaman, K.}, \bibinfo{author}{Loerke, D.}, \bibinfo{author}{Mettlen, M.}, \bibinfo{author}{Kuwata, H.}, \bibinfo{author}{Grinstein, S.}, \bibinfo{author}{Schmid, S.~L.}, and \bibinfo{author}{Danuser, G.} (\bibinfo{year}{2008}). \bibinfo{title}{Robust single-particle tracking in live-cell time-lapse sequences}.
\newblock \bibinfo{journal}{Nature Methods} \emph{\bibinfo{volume}{5}}, \bibinfo{pages}{695--702}. \DOIprefix\doi{10.1038/nmeth.1237}.
\bibitem[{O’Connor et~al.(2022)O’Connor, Alnahhas, Lugagne and Dunlop}]{oconnor_delta_2022}
\bibinfo{author}{O’Connor, O.~M.}, \bibinfo{author}{Alnahhas, R.~N.}, \bibinfo{author}{Lugagne, J.-B.}, and \bibinfo{author}{Dunlop, M.~J.} (\bibinfo{year}{2022}). \bibinfo{title}{{DeLTA} 2.0: {A} deep learning pipeline for quantifying single-cell spatial and temporal dynamics}.
\newblock \bibinfo{journal}{PLOS Computational Biology} \emph{\bibinfo{volume}{18}}, \bibinfo{pages}{e1009797}. \DOIprefix\doi{10.1371/journal.pcbi.1009797}.
\bibitem[{O'Connor and Dunlop(2025)}]{OConnor2025}
\bibinfo{author}{O'Connor, O.~M.}, and \bibinfo{author}{Dunlop, M.~J.} (\bibinfo{year}{2025}). \bibinfo{title}{{Cell-TRACTR: A transformer-based model for end-to-end segmentation and tracking of cells}}.
\newblock \bibinfo{journal}{PLOS Computational Biology} \emph{\bibinfo{volume}{21}}, \bibinfo{pages}{e1013071}. \DOIprefix\doi{10.1371/journal.pcbi.1013071}.
\bibitem[{Schwartz et~al.(2023)Schwartz, Moen, Miller, Dougherty, Borba, Ding, Graf, Pao and Valen}]{schwartz_caliban_2023}
\bibinfo{author}{Schwartz, M.~S.}, \bibinfo{author}{Moen, E.}, \bibinfo{author}{Miller, G.}, \bibinfo{author}{Dougherty, T.}, \bibinfo{author}{Borba, E.}, \bibinfo{author}{Ding, R.}, \bibinfo{author}{Graf, W.}, \bibinfo{author}{Pao, E.}, and \bibinfo{author}{Valen, D.~V.} (\bibinfo{year}{2023}).
\newblock \bibinfo{title}{Caliban: {Accurate} cell tracking and lineage construction in live-cell imaging experiments with deep learning}.
\newblock \bibinfo{publisher}{bioRxiv}.
\newblock \DOIprefix\doi{10.1101/803205}.
\bibitem[{Edlund et~al.(2021)Edlund, Jackson, Khalid, Bevan, Dale, Dengel, Ahmed, Trygg and Sjögren}]{edlund_livecelllarge-scale_2021}
\bibinfo{author}{Edlund, C.}, \bibinfo{author}{Jackson, T.~R.}, \bibinfo{author}{Khalid, N.}, \bibinfo{author}{Bevan, N.}, \bibinfo{author}{Dale, T.}, \bibinfo{author}{Dengel, A.}, \bibinfo{author}{Ahmed, S.}, \bibinfo{author}{Trygg, J.}, and \bibinfo{author}{Sjögren, R.} (\bibinfo{year}{2021}). \bibinfo{title}{{LIVECell} -- {A} large-scale dataset for label-free live cell segmentation}.
\newblock \bibinfo{journal}{Nature Methods} \emph{\bibinfo{volume}{18}}, \bibinfo{pages}{1038--1045}. \DOIprefix\doi{10.1038/s41592-021-01249-6}.
\bibitem[{Berg et~al.(2019)Berg, Kutra, Kroeger, Straehle, Kausler, Haubold, Schiegg, Ales, Beier, Rudy, Eren, Cervantes, Xu, Beuttenmueller, Wolny, Zhang, Koethe, Hamprecht and Kreshuk}]{berg_ilastik_2019}
\bibinfo{author}{Berg, S.}, \bibinfo{author}{Kutra, D.}, \bibinfo{author}{Kroeger, T.}, \bibinfo{author}{Straehle, C.~N.}, \bibinfo{author}{Kausler, B.~X.}, \bibinfo{author}{Haubold, C.}, \bibinfo{author}{Schiegg, M.}, \bibinfo{author}{Ales, J.}, \bibinfo{author}{Beier, T.}, \bibinfo{author}{Rudy, M.}, \bibinfo{author}{Eren, K.}, \bibinfo{author}{Cervantes, J.~I.}, \bibinfo{author}{Xu, B.}, \bibinfo{author}{Beuttenmueller, F.}, \bibinfo{author}{Wolny, A.}, \bibinfo{author}{Zhang, C.}, \bibinfo{author}{Koethe, U.}, \bibinfo{author}{Hamprecht, F.~A.}, and \bibinfo{author}{Kreshuk, A.} (\bibinfo{year}{2019}). \bibinfo{title}{ilastik: {I}nteractive machine learning for (bio)image analysis}.
\newblock \bibinfo{journal}{Nature Methods} \emph{\bibinfo{volume}{16}}, \bibinfo{pages}{1226--1232}. \DOIprefix\doi{10.1038/s41592-019-0582-9}.
\bibitem[{Stirling et~al.(2021)Stirling, Swain-Bowden, Lucas, Carpenter, Cimini and Goodman}]{stirling_cellprofiler_2021}
\bibinfo{author}{Stirling, D.~R.}, \bibinfo{author}{Swain-Bowden, M.~J.}, \bibinfo{author}{Lucas, A.~M.}, \bibinfo{author}{Carpenter, A.~E.}, \bibinfo{author}{Cimini, B.~A.}, and \bibinfo{author}{Goodman, A.} (\bibinfo{year}{2021}). \bibinfo{title}{{CellProfiler} 4: {I}mprovements in speed, utility and usability}.
\newblock \bibinfo{journal}{BMC Bioinformatics} \emph{\bibinfo{volume}{22}}, \bibinfo{pages}{433}. \DOIprefix\doi{10.1186/s12859-021-04344-9}.
\bibitem[{Stylianidou et~al.(2016)Stylianidou, Brennan, Nissen, Kuwada and Wiggins}]{stylianidou_supersegger_2016}
\bibinfo{author}{Stylianidou, S.}, \bibinfo{author}{Brennan, C.}, \bibinfo{author}{Nissen, S.~B.}, \bibinfo{author}{Kuwada, N.~J.}, and \bibinfo{author}{Wiggins, P.~A.} (\bibinfo{year}{2016}). \bibinfo{title}{{SuperSegger}: {R}obust image segmentation, analysis and lineage tracking of bacterial cells}.
\newblock \bibinfo{journal}{Molecular Microbiology} \emph{\bibinfo{volume}{102}}, \bibinfo{pages}{690--700}. \DOIprefix\doi{10.1111/mmi.13486}.
\bibitem[{Lo et~al.(2025)Lo, Cutler, {James Choi} and Wiggins}]{Lo2025}
\bibinfo{author}{Lo, T.~W.}, \bibinfo{author}{Cutler, K.~J.}, \bibinfo{author}{{James Choi}, H.}, and \bibinfo{author}{Wiggins, P.~A.} (\bibinfo{year}{2025}). \bibinfo{title}{{OmniSegger: A time-lapse image analysis pipeline for bacterial cells}}.
\newblock \bibinfo{journal}{PLOS Computational Biology} \emph{\bibinfo{volume}{21}}, \bibinfo{pages}{e1013088}. \DOIprefix\doi{10.1371/journal.pcbi.1013088}.
\bibitem[{Ouyang et~al.(2019)Ouyang, Mueller, Hjelmare, Lundberg and Zimmer}]{ouyang_imjoy_2019}
\bibinfo{author}{Ouyang, W.}, \bibinfo{author}{Mueller, F.}, \bibinfo{author}{Hjelmare, M.}, \bibinfo{author}{Lundberg, E.}, and \bibinfo{author}{Zimmer, C.} (\bibinfo{year}{2019}). \bibinfo{title}{{ImJoy}: {A}n open-source computational platform for the deep learning era}.
\newblock \bibinfo{journal}{Nature Methods} \emph{\bibinfo{volume}{16}}, \bibinfo{pages}{1199--1200}. \DOIprefix\doi{10.1038/s41592-019-0627-0}.
\bibitem[{Luik et~al.(2024)Luik, Rosas-Bertolini, Reits, Hoebe and Krawczyk}]{luik_biomero_2024}
\bibinfo{author}{Luik, T.~T.}, \bibinfo{author}{Rosas-Bertolini, R.}, \bibinfo{author}{Reits, E.~A.}, \bibinfo{author}{Hoebe, R.~A.}, and \bibinfo{author}{Krawczyk, P.~M.} (\bibinfo{year}{2024}). \bibinfo{title}{{BIOMERO: A scalable and extensible image analysis framework}}.
\newblock \bibinfo{journal}{Patterns} \emph{\bibinfo{volume}{5}}, \bibinfo{pages}{101024}. \DOIprefix\doi{10.1016/j.patter.2024.101024}.
\bibitem[{Schindelin et~al.(2012)Schindelin, Arganda-Carreras, Frise, Kaynig, Longair, Pietzsch, Preibisch, Rueden, Saalfeld, Schmid, Tinevez, White, Hartenstein, Eliceiri, Tomancak and Cardona}]{schindelin_fiji_2012}
\bibinfo{author}{Schindelin, J.}, \bibinfo{author}{Arganda-Carreras, I.}, \bibinfo{author}{Frise, E.}, \bibinfo{author}{Kaynig, V.}, \bibinfo{author}{Longair, M.}, \bibinfo{author}{Pietzsch, T.}, \bibinfo{author}{Preibisch, S.}, \bibinfo{author}{Rueden, C.}, \bibinfo{author}{Saalfeld, S.}, \bibinfo{author}{Schmid, B.}, \bibinfo{author}{Tinevez, J.-Y.}, \bibinfo{author}{White, D.~J.}, \bibinfo{author}{Hartenstein, V.}, \bibinfo{author}{Eliceiri, K.}, \bibinfo{author}{Tomancak, P.}, and \bibinfo{author}{Cardona, A.} (\bibinfo{year}{2012}). \bibinfo{title}{Fiji: {A}n open-source platform for biological-image analysis}.
\newblock \bibinfo{journal}{Nature Methods} \emph{\bibinfo{volume}{9}}, \bibinfo{pages}{676--682}. \DOIprefix\doi{10.1038/nmeth.2019}.
\bibitem[{Chiu et~al.(2022)Chiu, Clack and {the napari community}}]{chiu_napari_2022}
\bibinfo{author}{Chiu, C.-L.}, \bibinfo{author}{Clack, N.}, and \bibinfo{author}{{the napari community}} (\bibinfo{year}{2022}). \bibinfo{title}{napari: {A} {Python} multi-dimensional image viewer platform for the research community}.
\newblock \bibinfo{journal}{Microscopy and Microanalysis} \emph{\bibinfo{volume}{28}}, \bibinfo{pages}{1576--1577}. \DOIprefix\doi{10.1017/S1431927622006328}.
\bibitem[{Ershov et~al.(2022)Ershov, Phan, Pylvänäinen, Rigaud, Le~Blanc, Charles-Orszag, Conway, Laine, Roy, Bonazzi, Duménil, Jacquemet and Tinevez}]{ershov_trackmate_2022}
\bibinfo{author}{Ershov, D.}, \bibinfo{author}{Phan, M.-S.}, \bibinfo{author}{Pylvänäinen, J.~W.}, \bibinfo{author}{Rigaud, S.~U.}, \bibinfo{author}{Le~Blanc, L.}, \bibinfo{author}{Charles-Orszag, A.}, \bibinfo{author}{Conway, J. R.~W.}, \bibinfo{author}{Laine, R.~F.}, \bibinfo{author}{Roy, N.~H.}, \bibinfo{author}{Bonazzi, D.}, \bibinfo{author}{Duménil, G.}, \bibinfo{author}{Jacquemet, G.}, and \bibinfo{author}{Tinevez, J.-Y.} (\bibinfo{year}{2022}). \bibinfo{title}{{TrackMate} 7: {I}ntegrating state-of-the-art segmentation algorithms into tracking pipelines}.
\newblock \bibinfo{journal}{Nature Methods} \emph{\bibinfo{volume}{19}}, \bibinfo{pages}{829--832}. \DOIprefix\doi{10.1038/s41592-022-01507-1}.
\bibitem[{Ducret et~al.(2016)Ducret, Quardokus and Brun}]{ducret_microbej_2016}
\bibinfo{author}{Ducret, A.}, \bibinfo{author}{Quardokus, E.~M.}, and \bibinfo{author}{Brun, Y.~V.} (\bibinfo{year}{2016}). \bibinfo{title}{{MicrobeJ}, a tool for high throughput bacterial cell detection and quantitative analysis}.
\newblock \bibinfo{journal}{Nature Microbiology} \emph{\bibinfo{volume}{1}}, \bibinfo{pages}{1--7}. \DOIprefix\doi{10.1038/nmicrobiol.2016.77}.
\bibitem[{Ouyang et~al.(2022)Ouyang, Beuttenmueller, Gómez-de Mariscal, Pape, Burke, Garcia-López-de Haro, Russell, Moya-Sans, de-la Torre-Gutiérrez, Schmidt, Kutra, Novikov, Weigert, Schmidt, Bankhead, Jacquemet, Sage, Henriques, Muñoz-Barrutia, Lundberg, Jug and Kreshuk}]{ouyang_bioimage_2022}
\bibinfo{author}{Ouyang, W.}, \bibinfo{author}{Beuttenmueller, F.}, \bibinfo{author}{Gómez-de Mariscal, E.}, \bibinfo{author}{Pape, C.}, \bibinfo{author}{Burke, T.}, \bibinfo{author}{Garcia-López-de Haro, C.}, \bibinfo{author}{Russell, C.}, \bibinfo{author}{Moya-Sans, L.}, \bibinfo{author}{de-la Torre-Gutiérrez, C.}, \bibinfo{author}{Schmidt, D.}, \bibinfo{author}{Kutra, D.}, \bibinfo{author}{Novikov, M.}, \bibinfo{author}{Weigert, M.}, \bibinfo{author}{Schmidt, U.}, \bibinfo{author}{Bankhead, P.}, \bibinfo{author}{Jacquemet, G.}, \bibinfo{author}{Sage, D.}, \bibinfo{author}{Henriques, R.}, \bibinfo{author}{Muñoz-Barrutia, A.}, \bibinfo{author}{Lundberg, E.}, \bibinfo{author}{Jug, F.}, and \bibinfo{author}{Kreshuk, A.} (\bibinfo{year}{2022}).
\newblock \bibinfo{title}{{BioImage} {Model} {Zoo}: {A} community-driven resource for accessible {Deep} {Learning} in bioimage analysis}.
\newblock \bibinfo{publisher}{bioRxiv}.
\newblock \DOIprefix\doi{10.1101/2022.06.07.495102}.
\bibitem[{Kluyver et~al.(2016)Kluyver, Ragan-Kelley, P{\'e}rez, Rez, Granger, Bussonnier, Frederic, Kelley, Hamrick, Grout, Corlay, Ivanov, Avila, Abdalla, Willing and Team}]{kluyver_jupyter_2016}
\bibinfo{author}{Kluyver, T.}, \bibinfo{author}{Ragan-Kelley, B.}, \bibinfo{author}{P{\'e}rez}, \bibinfo{author}{Rez, F.}, \bibinfo{author}{Granger, B.}, \bibinfo{author}{Bussonnier, M.}, \bibinfo{author}{Frederic, J.}, \bibinfo{author}{Kelley, K.}, \bibinfo{author}{Hamrick, J.}, \bibinfo{author}{Grout, J.}, \bibinfo{author}{Corlay, S.}, \bibinfo{author}{Ivanov, P.}, \bibinfo{author}{Avila, D.}, \bibinfo{author}{Abdalla, S.}, \bibinfo{author}{Willing, C.}, and \bibinfo{author}{Team, J.~D.}
\newblock \bibinfo{title}{Jupyter {Notebooks} – a publishing format for reproducible computational workflows}.
\newblock In: \bibinfo{booktitle}{Positioning and {Power} in {Academic} {Publishing}: {Players}, {Agents} and {Agendas}} ( \bibinfo{pages}{87--90}). \bibinfo{publisher}{IOS Press} (\bibinfo{year}{2016}):\unskip( \bibinfo{pages}{87--90}).
\newblock \DOIprefix\doi{10.3233/978-1-61499-649-1-87}.
\bibitem[{von Chamier et~al.(2021)von Chamier, Laine, Jukkala, Spahn, Krentzel, Nehme, Lerche, Hernández-Pérez, Mattila, Karinou, Holden, Solak, Krull, Buchholz, Jones, Royer, Leterrier, Shechtman, Jug, Heilemann, Jacquemet and Henriques}]{von_chamier_democratising_2021}
\bibinfo{author}{von Chamier, L.}, \bibinfo{author}{Laine, R.~F.}, \bibinfo{author}{Jukkala, J.}, \bibinfo{author}{Spahn, C.}, \bibinfo{author}{Krentzel, D.}, \bibinfo{author}{Nehme, E.}, \bibinfo{author}{Lerche, M.}, \bibinfo{author}{Hernández-Pérez, S.}, \bibinfo{author}{Mattila, P.~K.}, \bibinfo{author}{Karinou, E.}, \bibinfo{author}{Holden, S.}, \bibinfo{author}{Solak, A.~C.}, \bibinfo{author}{Krull, A.}, \bibinfo{author}{Buchholz, T.-O.}, \bibinfo{author}{Jones, M.~L.}, \bibinfo{author}{Royer, L.~A.}, \bibinfo{author}{Leterrier, C.}, \bibinfo{author}{Shechtman, Y.}, \bibinfo{author}{Jug, F.}, \bibinfo{author}{Heilemann, M.}, \bibinfo{author}{Jacquemet, G.}, and \bibinfo{author}{Henriques, R.} (\bibinfo{year}{2021}). \bibinfo{title}{Democratising deep learning for microscopy with {ZeroCostDL4Mic}}.
\newblock \bibinfo{journal}{Nature Communications} \emph{\bibinfo{volume}{12}}, \bibinfo{pages}{2276}. \DOIprefix\doi{10.1038/s41467-021-22518-0}.
\bibitem[{Harris et~al.(2020)Harris, Millman, van~der Walt, Gommers, Virtanen, Cournapeau, Wieser, Taylor, Berg, Smith, Kern, Picus, Hoyer, van Kerkwijk, Brett, Haldane, del Río, Wiebe, Peterson, Gérard-Marchant, Sheppard, Reddy, Weckesser, Abbasi, Gohlke and Oliphant}]{harris_array_2020}
\bibinfo{author}{Harris, C.~R.}, \bibinfo{author}{Millman, K.~J.}, \bibinfo{author}{van~der Walt, S.~J.}, \bibinfo{author}{Gommers, R.}, \bibinfo{author}{Virtanen, P.}, \bibinfo{author}{Cournapeau, D.}, \bibinfo{author}{Wieser, E.}, \bibinfo{author}{Taylor, J.}, \bibinfo{author}{Berg, S.}, \bibinfo{author}{Smith, N.~J.}, \bibinfo{author}{Kern, R.}, \bibinfo{author}{Picus, M.}, \bibinfo{author}{Hoyer, S.}, \bibinfo{author}{van Kerkwijk, M.~H.}, \bibinfo{author}{Brett, M.}, \bibinfo{author}{Haldane, A.}, \bibinfo{author}{del Río, J.~F.}, \bibinfo{author}{Wiebe, M.}, \bibinfo{author}{Peterson, P.}, \bibinfo{author}{Gérard-Marchant, P.}, \bibinfo{author}{Sheppard, K.}, \bibinfo{author}{Reddy, T.}, \bibinfo{author}{Weckesser, W.}, \bibinfo{author}{Abbasi, H.}, \bibinfo{author}{Gohlke, C.}, and \bibinfo{author}{Oliphant, T.~E.} (\bibinfo{year}{2020}). \bibinfo{title}{Array programming with {NumPy}}.
\newblock \bibinfo{journal}{Nature} \emph{\bibinfo{volume}{585}}, \bibinfo{pages}{357--362}. \DOIprefix\doi{10.1038/s41586-020-2649-2}.
\bibitem[{Allan et~al.(2012)Allan, Burel, Moore, Blackburn, Linkert, Loynton, MacDonald, Moore, Neves, Patterson, Porter, Tarkowska, Loranger, Avondo, Lagerstedt, Lianas, Leo, Hands, Hay, Patwardhan, Best, Kleywegt, Zanetti and Swedlow}]{allan_omero_2012}
\bibinfo{author}{Allan, C.}, \bibinfo{author}{Burel, J.-M.}, \bibinfo{author}{Moore, J.}, \bibinfo{author}{Blackburn, C.}, \bibinfo{author}{Linkert, M.}, \bibinfo{author}{Loynton, S.}, \bibinfo{author}{MacDonald, D.}, \bibinfo{author}{Moore, W.~J.}, \bibinfo{author}{Neves, C.}, \bibinfo{author}{Patterson, A.}, \bibinfo{author}{Porter, M.}, \bibinfo{author}{Tarkowska, A.}, \bibinfo{author}{Loranger, B.}, \bibinfo{author}{Avondo, J.}, \bibinfo{author}{Lagerstedt, I.}, \bibinfo{author}{Lianas, L.}, \bibinfo{author}{Leo, S.}, \bibinfo{author}{Hands, K.}, \bibinfo{author}{Hay, R.~T.}, \bibinfo{author}{Patwardhan, A.}, \bibinfo{author}{Best, C.}, \bibinfo{author}{Kleywegt, G.~J.}, \bibinfo{author}{Zanetti, G.}, and \bibinfo{author}{Swedlow, J.~R.} (\bibinfo{year}{2012}). \bibinfo{title}{{OMERO}: {F}lexible, model-driven data management for experimental biology}.
\newblock \bibinfo{journal}{Nature Methods} \emph{\bibinfo{volume}{9}}, \bibinfo{pages}{245--253}. \DOIprefix\doi{10.1038/nmeth.1896}.
\bibitem[{Pachitariu et~al.(2025)Pachitariu, Rariden and Stringer}]{pachitariu_cellpose-sam_2025}
\bibinfo{author}{Pachitariu, M.}, \bibinfo{author}{Rariden, M.}, and \bibinfo{author}{Stringer, C.} (\bibinfo{year}{2025}).
\newblock \bibinfo{title}{Cellpose-{SAM}: {S}uperhuman generalization for cellular segmentation}.
\newblock \bibinfo{publisher}{bioRxiv}.
\newblock \DOIprefix\doi{10.1101/2025.04.28.651001}.
\bibitem[{Paul et~al.(2025)Paul, Seiffarth, R{\"{u}}gamer, Scharr and N{\"{o}}h}]{paul2025}
\bibinfo{author}{Paul, R.~D.}, \bibinfo{author}{Seiffarth, J.}, \bibinfo{author}{R{\"{u}}gamer, D.}, \bibinfo{author}{Scharr, H.}, and \bibinfo{author}{N{\"{o}}h, K.} (\bibinfo{year}{2025}). \bibinfo{title}{How to make your cell tracker say {"I dunno!"}}.
\newblock \bibinfo{journal}{arXiv [CS]}. \DOIprefix\doi{10.48550/arXiv.2503.09244}.
\bibitem[{Bragantini et~al.(2025)Bragantini, Lange and Royer}]{bragantini_large-scale_2025}
\bibinfo{author}{Bragantini, J.}, \bibinfo{author}{Lange, M.}, and \bibinfo{author}{Royer, L.} (\bibinfo{year}{2025}).
\newblock \bibinfo{title}{Large-scale multi-hypotheses cell tracking using ultrametric contours maps}.
\newblock In: \bibinfo{editor}{Leonardis, A.}, \bibinfo{editor}{Ricci, E.}, \bibinfo{editor}{Roth, S.}, \bibinfo{editor}{Russakovsky, O.}, \bibinfo{editor}{Sattler, T.}, and \bibinfo{editor}{Varol, G.}, eds. \bibinfo{booktitle}{Computer {Vision} – {ECCV} 2024}. \bibinfo{address}{Cham}: \bibinfo{publisher}{Springer Nature Switzerland} ( \bibinfo{pages}{36--54}).
\newblock \DOIprefix\doi{10.1007/978-3-031-72986-7_3}.
\bibitem[{Seiffarth and Nöh(2025)}]{Seiffarth_2025}
\bibinfo{author}{Seiffarth, J.}, and \bibinfo{author}{Nöh, K.} (\bibinfo{year}{2025}). \bibinfo{title}{{PyUAT: Open-source Python framework for efficient and scalable cell tracking}}.
\newblock \bibinfo{journal}{arXiv [q-bio]}. \DOIprefix\doi{10.48550/arXiv.2503.21914}.
\bibitem[{{The Pandas development team}(2024)}]{team_pandas-devpandas_2024}
\bibinfo{author}{{The Pandas development team}} (\bibinfo{year}{2024}).
\newblock \bibinfo{title}{pandas-dev/pandas: {Pandas}}.
\newblock \bibinfo{publisher}{Zenodo}.
\newblock \DOIprefix\doi{10.5281/zenodo.13819579}.
\bibitem[{Hunter(2007)}]{hunter_matplotlib_2007}
\bibinfo{author}{Hunter, J.~D.} (\bibinfo{year}{2007}). \bibinfo{title}{Matplotlib: {A} {2D} graphics environment}.
\newblock \bibinfo{journal}{Computing in Science \& Engineering} \emph{\bibinfo{volume}{9}}, \bibinfo{pages}{90--95}. \DOIprefix\doi{10.1109/MCSE.2007.55}.
\bibitem[{Waskom(2021)}]{waskom_seaborn_2021}
\bibinfo{author}{Waskom, M.~L.} (\bibinfo{year}{2021}). \bibinfo{title}{seaborn: {S}tatistical data visualization}.
\newblock \bibinfo{journal}{Journal of Open Source Software} \emph{\bibinfo{volume}{6}}, \bibinfo{pages}{3021}. \DOIprefix\doi{10.21105/joss.03021}.
\bibitem[{Seiffarth et~al.(2024)Seiffarth, Blöbaum, Grünberger and Nöh}]{seiffarth_customizable_2024}
\bibinfo{author}{Seiffarth, J.}, \bibinfo{author}{Blöbaum, L.}, \bibinfo{author}{Grünberger, A.}, and \bibinfo{author}{Nöh, K.} (\bibinfo{year}{2024}).
\newblock \bibinfo{title}{Customizable and interactive visualizations for investigating spatio-temporal single-cell information}.
\newblock In: \bibinfo{booktitle}{2024 {IEEE} {International} {Symposium} on {Biomedical} {Imaging} ({ISBI})}. ( \bibinfo{pages}{1--5}).
\newblock \DOIprefix\doi{10.1109/ISBI56570.2024.10635297}.
\bibitem[{Grünberger et~al.(2013)Grünberger, Ooyen, Paczia, Rohe, Schiendzielorz, Eggeling, Wiechert, Kohlheyer and Noack}]{grunberger_beyond_2013}
\bibinfo{author}{Grünberger, A.}, \bibinfo{author}{Ooyen, J.~v.}, \bibinfo{author}{Paczia, N.}, \bibinfo{author}{Rohe, P.}, \bibinfo{author}{Schiendzielorz, G.}, \bibinfo{author}{Eggeling, L.}, \bibinfo{author}{Wiechert, W.}, \bibinfo{author}{Kohlheyer, D.}, and \bibinfo{author}{Noack, S.} (\bibinfo{year}{2013}). \bibinfo{title}{Beyond growth rate 0.6: \textit{{C}orynebacterium glutamicum} cultivated in highly diluted environments}.
\newblock \bibinfo{journal}{Biotechnology and Bioengineering} \emph{\bibinfo{volume}{110}}, \bibinfo{pages}{220--228}. \DOIprefix\doi{10.1002/bit.24616}.
\bibitem[{Schmitz et~al.(2024)Schmitz, Yermakov and Grünberger}]{schmitz_protocol_2024}
\bibinfo{author}{Schmitz, J.}, \bibinfo{author}{Yermakov, B.}, and \bibinfo{author}{Grünberger, A.} (\bibinfo{year}{2024}). \bibinfo{title}{Protocol for microfluidic single-cell cultivation and live-cell imaging of {Chinese} hamster ovary suspension cell lines}.
\newblock \bibinfo{journal}{STAR Protocols} \emph{\bibinfo{volume}{5}}, \bibinfo{pages}{103106}. \DOIprefix\doi{10.1016/j.xpro.2024.103106}.
\bibitem[{Kendall(1948)}]{kendall_role_1948}
\bibinfo{author}{Kendall, D.~G.} (\bibinfo{year}{1948}). \bibinfo{title}{On the role of variable generation time in the development of a stochastic birth process}.
\newblock \bibinfo{journal}{Biometrika} \emph{\bibinfo{volume}{35}}, \bibinfo{pages}{316}. \DOIprefix\doi{10.2307/2332354}.
\bibitem[{Yates et~al.(2017)Yates, Ford and Mort}]{yates_multi-stage_2017}
\bibinfo{author}{Yates, C.~A.}, \bibinfo{author}{Ford, M.~J.}, and \bibinfo{author}{Mort, R.~L.} (\bibinfo{year}{2017}). \bibinfo{title}{A multi-stage representation of cell proliferation as a {Markov} process}.
\newblock \bibinfo{journal}{Bulletin of Mathematical Biology} \emph{\bibinfo{volume}{79}}, \bibinfo{pages}{2905--2928}. \DOIprefix\doi{10.1007/s11538-017-0356-4}.
\bibitem[{Paul et~al.(2024)Paul, Seiffarth, Scharr and Nöh}]{paul_robust_2024}
\bibinfo{author}{Paul, R.~D.}, \bibinfo{author}{Seiffarth, J.}, \bibinfo{author}{Scharr, H.}, and \bibinfo{author}{Nöh, K.} (\bibinfo{year}{2024}).
\newblock \bibinfo{title}{Robust approximate characterization of single-cell heterogeneity in microbial growth}.
\newblock In: \bibinfo{booktitle}{2024 {IEEE} {International} {Symposium} on {Biomedical} {Imaging} ({ISBI})}. \bibinfo{address}{Athens, Greece}: \bibinfo{publisher}{IEEE} ( \bibinfo{pages}{1--5}).
\newblock \DOIprefix\doi{10.1109/ISBI56570.2024.10635267}.
\bibitem[{Hein and Jafarpour(2024)}]{hein_competition_2024}
\bibinfo{author}{Hein, Y.}, and \bibinfo{author}{Jafarpour, F.} (\bibinfo{year}{2024}). \bibinfo{title}{Competition between transient oscillations and early stochasticity in exponentially growing populations}.
\newblock \bibinfo{journal}{Physical Review Research} \emph{\bibinfo{volume}{6}}, \bibinfo{pages}{033320}. \DOIprefix\doi{10.1103/PhysRevResearch.6.033320}.
\bibitem[{Virtanen et~al.(2020)Virtanen, Gommers, Oliphant, Haberland, Reddy, Cournapeau, Burovski, Peterson, Weckesser, Bright, van~der Walt, Brett, Wilson, Millman, Mayorov, Nelson, Jones, Kern, Larson, Carey, Polat, Feng, Moore, VanderPlas, Laxalde, Perktold, Cimrman, Henriksen, Quintero, Harris, Archibald, Ribeiro, Pedregosa, van Mulbregt and {SciPy 1.0 Contributors}}]{virtanen_scipy_2020}
\bibinfo{author}{Virtanen, P.}, \bibinfo{author}{Gommers, R.}, \bibinfo{author}{Oliphant, T.~E.}, \bibinfo{author}{Haberland, M.}, \bibinfo{author}{Reddy, T.}, \bibinfo{author}{Cournapeau, D.}, \bibinfo{author}{Burovski, E.}, \bibinfo{author}{Peterson, P.}, \bibinfo{author}{Weckesser, W.}, \bibinfo{author}{Bright, J.}, \bibinfo{author}{van~der Walt, S.~J.}, \bibinfo{author}{Brett, M.}, \bibinfo{author}{Wilson, J.}, \bibinfo{author}{Millman, K.~J.}, \bibinfo{author}{Mayorov, N.}, \bibinfo{author}{Nelson, A. R.~J.}, \bibinfo{author}{Jones, E.}, \bibinfo{author}{Kern, R.}, \bibinfo{author}{Larson, E.}, \bibinfo{author}{Carey, C.~J.}, \bibinfo{author}{Polat, I.}, \bibinfo{author}{Feng, Y.}, \bibinfo{author}{Moore, E.~W.}, \bibinfo{author}{VanderPlas, J.}, \bibinfo{author}{Laxalde, D.}, \bibinfo{author}{Perktold, J.}, \bibinfo{author}{Cimrman, R.}, \bibinfo{author}{Henriksen, I.}, \bibinfo{author}{Quintero, E.~A.}, \bibinfo{author}{Harris, C.~R.}, \bibinfo{author}{Archibald, A.~M.}, \bibinfo{author}{Ribeiro,
  A.~H.}, \bibinfo{author}{Pedregosa, F.}, \bibinfo{author}{van Mulbregt, P.}, and \bibinfo{author}{{SciPy 1.0 Contributors}} (\bibinfo{year}{2020}). \bibinfo{title}{{SciPy} 1.0: {F}undamental algorithms for scientific computing in {Python}}.
\newblock \bibinfo{journal}{Nature Methods} \emph{\bibinfo{volume}{17}}, \bibinfo{pages}{261--272}. \DOIprefix\doi{10.1038/s41592-019-0686-2}.
\bibitem[{Tsien(1998)}]{tsien_green_1998}
\bibinfo{author}{Tsien, R.~Y.} (\bibinfo{year}{1998}). \bibinfo{title}{The green fluorescent protein}.
\newblock \bibinfo{journal}{Annual Review of Biochemistry} \emph{\bibinfo{volume}{67}}, \bibinfo{pages}{509--544}. \DOIprefix\doi{10.1146/annurev.biochem.67.1.509}.
\bibitem[{Royer(2024)}]{royer_omega_2024}
\bibinfo{author}{Royer, L.~A.} (\bibinfo{year}{2024}). \bibinfo{title}{Omega -- harnessing the power of large language models for bioimage analysis}.
\newblock \bibinfo{journal}{Nature Methods} \emph{\bibinfo{volume}{21}}, \bibinfo{pages}{1371--1373}. \DOIprefix\doi{10.1038/s41592-024-02310-w}.
\bibitem[{Weil et~al.(2023)Weil, Schneider, Tschöpe, Bauer, Maus, Frey, Brilhaus, Martins~Rodrigues, Doniparthi, Wetzels, Lukasczyk, Kranz, Grüning, Zimmer, Deßloch, von Suchodoletz, Usadel, Garth and Mühlhaus}]{weil_plantdatahub_2023}
\bibinfo{author}{Weil, H.~L.}, \bibinfo{author}{Schneider, K.}, \bibinfo{author}{Tschöpe, M.}, \bibinfo{author}{Bauer, J.}, \bibinfo{author}{Maus, O.}, \bibinfo{author}{Frey, K.}, \bibinfo{author}{Brilhaus, D.}, \bibinfo{author}{Martins~Rodrigues, C.}, \bibinfo{author}{Doniparthi, G.}, \bibinfo{author}{Wetzels, F.}, \bibinfo{author}{Lukasczyk, J.}, \bibinfo{author}{Kranz, A.}, \bibinfo{author}{Grüning, B.}, \bibinfo{author}{Zimmer, D.}, \bibinfo{author}{Deßloch, S.}, \bibinfo{author}{von Suchodoletz, D.}, \bibinfo{author}{Usadel, B.}, \bibinfo{author}{Garth, C.}, and \bibinfo{author}{Mühlhaus, T.} (\bibinfo{year}{2023}). \bibinfo{title}{{PLANTdataHUB}: {A} collaborative platform for continuous {FAIR} data sharing in plant research}.
\newblock \bibinfo{journal}{The Plant Journal} \emph{\bibinfo{volume}{116}}, \bibinfo{pages}{974--988}. \DOIprefix\doi{10.1111/tpj.16474}.

\end{thebibliography}

\clearpage



\section*{Supplementary material}

# S.1 Strain, plasmid and oligonucleotide details

*Table 1: Bacterial strains used in this study*

| Strain | Relevant characteristics | Reference |
|---|---|---|
| ***E. coli*** | | |
| DH5α | F⁻ φ80*dlac* Δ(*lacZ*)M15 Δ(*lacZYA-argF*) U169 *endA1 recA1 hsdR17* (r$_K^-$ m$_K^+$) *deoR thi-1 phoA supE44* λ⁻ *gyrA96 relA1*; strain used for cloning procedures | Hanahan, 1983 |
| ***C. glutamicum*** | | |
| ATCC13032 (WT) | Wild type, biotin auxotrophic | Kinoshita et al. 1957 |
| ATCC13032::P*$_{odhI}$*-*e2-crimson* | Wild type derivative with an insertion of the gene *e2-crimson* under the control of the *odhI*-promotor in the intergenic region between cg1121-cg1122 | This study |
| ATCC13032::P*$_{odhI}$*-*mvenus* | Wild type derivative with an insertion of the gene *mvenus* under the control of the *odhI*-promotor in the intergenic region cg1121-cg1122 | This study |

*Table 2: Plasmids used in this study*

| Plasmid | Relevant characteristics | Source or reference |
|---|---|---|
| pK18*mobsacB*-int-P$_{tac}$-*crimson* | Kan$^r$; pK18*mobsacB* derivative carrying an insertion of the gene *e2-crimson* under control of the *tac* promotor | Baumgart et al. 2013 |
| pPREx2-*mvenus* | Kan$^r$; pPREx2 derivative carrying the mVenus encoding sequence under the control of the *tac* promoter | Sundermeyer et al. 2023 |
| pK18*mobsacB*-P*$_{odhI}$*-*e2-crimson* | Kan$^r$; pK18*mobsacB* derivative carrying an insertion of the gene *e2-crimson* under the control of the *odhI* promotor | This study |
| pK18*mobsacB*-P*$_{odhI}$*-*mvenus* | Kan$^r$; pK18*mobsacB* derivative carrying an insertion of the gene *mvenus* under the control of the *odhI* promotor | This study |

*Table 3: Oligonucleotides used in this study*

| Name | Sequence |
| --- | --- |
| Construction of pK18*mobsacB*-P*odhI*-*e2-crimson* | |
| pK18mobsacB_PodhI fw | ATTCCTCTTGCTCGTGTCTTGGCGCGTCCATCAGCAAC |
| PodhI_e2-crimson rv | GTTCTCAGTGCTATCCATTAAACTTCCTCCGTGTCG |
| e2-crimson fw | ATGGATAGCACTGAGAACG |
| e2-crimson_pK18mobsacB rv | TTGTGTCCATGAGTTCGCCTACTGGAACAGGTGGTG |
| Construction of pK18*mobsacB*-P*odhI*-*mvenus* | |
| pK18mobsacB_PodhI fw | ATTCCTCTTGCTCGTGTCTTGGCGCGTCCATCAGCAAC |
| PodhI_mvenus rv | TCTCCTTTGCTAGCCATTAAACTTCCTCCGTGTCGAC |
| mvenus fw | ATGGCTAGCAAAGGAGAAGAAC |
| mvenus_pK18mobsacB rv | TTGTGTCCATGAGTTCGCTTATTTGTAGAGCTCATCCATGCC |
| Confirmation of integration by colony-PCR | |
| IGR cg1121-22 fw | CTTGGTTCGAATATGCAGTTCGG |
| IGR cg1121-22 rv | AGCGTAAGGCCCCTACTTCCTG |

# Supplementary References

1. Hanahan D. (1983). Studies on transformation of *Escherichia coli* with plasmids. Journal of Molecular Biology *166*, 557–80. doi: 10.1016/s0022-2836(83)80284-8

2. Kinoshita, S., Udaka, S. and Shimono, M. (1957) Studies on amino scid fermentation production of L-glutamic acid by various microorganisms. Journal of General and Applied Microbiology *3*, 193–205. doi: 10.2323/jgam.3.193

3. Baumgart M, Unthan S, Rückert C, Sivalingam J, Grünberger A, Kalinowski J, Bott M, Noack S, Frunzke J. 2013. Construction of a prophage-free variant of *Corynebacterium glutamicum* ATCC 13032 for use as a platform strain for basic research and industrial biotechnology. (2013) Applied Environmental Microbiology *79*, 6006–15. doi: 10.1128/AEM.01634-13

4. Sundermeyer L, Folkerts J, Lückel B, Mack C, Baumgart M and Bott M. (2023) Cellular localization of the hybrid pyruvate/2-oxoglutarate dehydrogenase complex in the actinobacterium *Corynebacterium glutamicum* . Micrology Spectrum *11*, e02668-23. doi: 10.1128/spectrum.02668-23



\clearpage
\section*{S.2 Microfluidic Co-culture Cultivation}

We utilize two \CG{} strains with chromosomally integrated fluorescence markers under the control of the \textit{odhI} promoter, encoding either the \mvenus{} (blue) or \crimson{} (red) fluorescent protein (see SI~1). The two strains were cultured on an agar plate for 48 hours at 30$^\circ$C. This was followed by a three-stage liquid cultivation in 15 ml of medium each in a baffled flask at 30$^\circ$C for a total of 28~hours and 120~rpm shaking frequency. Brain Heart Infusion Medium (BHI) was used for the first stage. For the second and third stage, CGXII was used with different concentrations of iron sulphate (FeSO\textsubscript{4}), protocatechuic acid (PCA) and glucose. The cells were introduced into the chambers of the chip using the rapid inoculation method described by Probst et al. (2015). The optical density (OD600) of the cell suspension used to inoculate the chip was 0.5.\\[5em]

\subsection*{Supplementary References}
\begin{enumerate}
    \item Probst, C., Grünberger, A., Braun, N., Helfrich, S., Nöh, K., Wiechert, W., and Kohlheyer, D. (2015). Rapid inoculation of single bacteria into parallel picoliter fermentation chambers. Analalytical Methods 7, 91–98. doi:\href{https://doi.org/10.1039/C4AY02257B}{10.1039/C4AY02257B}
\end{enumerate}

\clearpage
\section*{S.3 Computing instantaneous growth rates of individual cells}

Let $a_1, \ldots, a_T$ be the measured single-cell area at time points $1, \ldots, T$. We define the instantaneous growth rate (IGR) $\mu_t$ of a cell at time $t$

\begin{equation}
    \mu_{\Delta_t} = \frac{a_{t+1} - a_{t}}{\Delta_t}
\end{equation}
where $\Delta_t$ denotes the time difference between the recorded images at time points $t$ and $t+1$. To remove the noise from the measurements, we apply scipy's gaussian filter to the extracted instantaneous growth rate time-series with a standard deviation $\sigma=4$~(Virtanen et al. 2020).\\[5em]

\subsection*{Supplementary References}
\begin{enumerate}
    \item Virtanen, P., Gommers, R., Oliphant, T. E., Haberland, M., Reddy, T., Cournapeau, D., Burovski, E., Peterson, P., Weckesser, W., Bright, J., van der Walt, S. J., Brett, M., Wilson, J., Millman, K. J., Mayorov, N., Nelson, A. R. J., Jones, E., Kern, R., Larson, E., Carey, C. J., Polat, I., Feng, Y., Moore, E. W., VanderPlas, J., Laxalde, D., Perktold, J., Cimrman, R., Henriksen, I., Quintero, E. A., Harris, C. R., Archibald, A. M., Ribeiro, A. H., Pedregosa, F., van Mulbregt, P., and SciPy 1.0 Contributors (2020). SciPy 1.0: Fundamental algorithms for scientific computing in Python. Nature Methods 17, 261–272.\href{https://doi.org/doi:10.1038/s41592-019-0686-2}{doi:10.1038/s41592-019-0686-2}
\end{enumerate}

\end{document}